\documentclass[journal]{IEEEtran}

\usepackage{times}
\usepackage{epsfig}
\usepackage{graphicx}
\usepackage{amsmath}
\usepackage{amssymb}
\usepackage{url}
\usepackage{amsfonts}
\usepackage{xcolor}
\usepackage{bm}
\usepackage{multirow}
\usepackage{colortbl}
\usepackage{hhline}
\usepackage{pifont}
\usepackage{mathtools}
\usepackage{wrapfig}
\usepackage{cite}
\usepackage[caption=false]{subfig}
\usepackage{ragged2e}
\usepackage{makecell}
\usepackage{enumitem}
\usepackage{wasysym}  
\usepackage{threeparttable}
\usepackage{booktabs}
\usepackage{siunitx}
\usepackage{array}
\usepackage{tikz}
\usepackage[T1]{fontenc}
\usepackage{microtype} 
 \usepackage{mathtools} 

\usetikzlibrary{
    matrix,
    positioning, 
    arrows.meta, 
    backgrounds, 
    fit, 
    calc
}

\definecolor{intro_fill}{HTML}{E9F2FF}
\definecolor{intro_border}{HTML}{5B8FD9}

\definecolor{bg_fill}{HTML}{E6F6FF}
\definecolor{bg_border}{HTML}{4B9ED9}

\definecolor{tax_fill}{HTML}{E0F7FD}
\definecolor{tax_border}{HTML}{39B4D1}

\definecolor{data_fill}{HTML}{DAFAF6}
\definecolor{data_border}{HTML}{32BEBA}

\definecolor{perf_fill}{HTML}{D9F7EE}
\definecolor{perf_border}{HTML}{37C4A3}

\definecolor{chall_fill}{HTML}{DEF6EA}
\definecolor{chall_border}{HTML}{41BA87}

\definecolor{con_fill}{HTML}{E6F5EA}
\definecolor{con_border}{HTML}{3FA86F}

\definecolor{grad_start}{HTML}{E8F2FF} 
\definecolor{grad_end}{HTML}{EAF7F0}   

\definecolor{frame_border_color}{HTML}{8DBCE6}

\definecolor{wmPurple}{HTML}{7A5EA4}

\usepackage[pagebackref=true,breaklinks=true,colorlinks,bookmarks=false]{hyperref}
\hypersetup{colorlinks=true, linkcolor=blue}

\graphicspath{{./Imgs/}}
\DeclareGraphicsExtensions{.pdf,.jpg,.png}

\newcommand{\cmark}{\ding{51}}

\newcommand{\rot}[1]{\rotatebox[origin=c]{90}{\strut #1}}
\newcolumntype{C}[1]{>{\centering\arraybackslash}p{#1}}

\newcommand{\para}[1]{\vspace{.05in}\noindent\textbf{#1}\quad}
\definecolor{mygray}{gray}{.92}

\newcommand{\figref}[1]{Fig.~\ref{#1}}
\newcommand{\tabref}[1]{Tab.~\ref{#1}}
\newcommand{\secref}[1]{\S\ref{#1}}

\newcommand{\equref}[1]{Eq.~(\ref{#1})}

\def\eg{\emph{e.g.}}
\def\vs{\emph{vs.~}}
\def\etc{\emph{etc}}
\def\etal{{\em et al.~}}

\begin{document}

\title{A Comprehensive Survey on World Models for Embodied AI}


\author{
  Xinqing Li, Xin He, 
  Le Zhang,~\IEEEmembership{Member,~IEEE}, 
  Min Wu,~\IEEEmembership{Senior Member,~IEEE},\\
  Xiaoli Li,~\IEEEmembership{Fellow,~IEEE}, 
  and Yun Liu
  \IEEEcompsocitemizethanks{%
    \IEEEcompsocthanksitem Xinqing Li and Yun Liu are with the College of Computer Science and the Academy for Advanced Interdisciplinary Studies, Nankai University, Tianjin 300350, China (e-mail: lixinqing@mail.nankai.edu.cn; liuyun@nankai.edu.cn).
    \IEEEcompsocthanksitem Xin He is with the School of Computer Science and Engineering, Tianjin University of Technology, Tianjin 300384, China (e-mail: hexin@email.tjut.edu.cn).
    \IEEEcompsocthanksitem Le Zhang is with the School of Information and Communication Engineering, University of Electronic Science and Technology of China, Chengdu 611731, Sichuan, China (e-mail: lezhang@uestc.edu.cn).
    \IEEEcompsocthanksitem Min Wu is with the Institute for Infocomm Research (I2R), Agency for Science, Technology and Research (A*STAR), 138632, Singapore (e-mail: wumin@i2r.a-star.edu.sg).
    \IEEEcompsocthanksitem Xiaoli Li is with the Information Systems Technology and Design (ISTD) Pillar, Singapore University of Technology and Design (SUTD), 487372, Singapore (e-mail: xiaoli\_li@sutd.edu.sg).
    \IEEEcompsocthanksitem Corresponding author: Yun Liu (e-mail: liuyun@nankai.edu.cn)
    \IEEEcompsocthanksitem This work was supported in part by the National Natural Science Foundation of China (No. 62576176) and in part by the Fundamental Research Funds for the Central Universities (Nankai University, No. 070-63253235).
  }
}

\IEEEtitleabstractindextext{%
\begin{abstract} \justifying
Embodied AI requires agents that perceive, act, and anticipate how actions reshape future world states. World models serve as internal simulators that capture environment dynamics, enabling forward and counterfactual rollouts to support perception, prediction, and decision making. This survey presents a unified framework for world models in embodied AI. Specifically, we formalize the problem setting and learning objectives, and propose a three-axis taxonomy encompassing: (1) \textbf{Functionality}, \emph{Decision-Coupled} \vs \emph{General-Purpose}; (2) \textbf{Temporal Modeling}, \emph{Sequential Simulation and Inference} \vs \emph{Global Difference Prediction}; (3) \textbf{Spatial Representation}, \emph{Global Latent Vector}, \emph{Token Feature Sequence}, \emph{Spatial Latent Grid}, and \emph{Decomposed Rendering Representation}. We systematize data resources and metrics across robotics, autonomous driving, and general video settings, covering pixel prediction quality, state-level understanding, and task performance. Furthermore, we offer a quantitative comparison of state-of-the-art models and distill key open challenges, including the scarcity of unified datasets and the need for evaluation metrics that assess physical consistency over pixel fidelity, the trade-off between model performance and the computational efficiency required for real-time control, and the core modeling difficulty of achieving long-horizon temporal consistency while mitigating error accumulation. Finally, we maintain a curated bibliography at \href{https://github.com/Li-Zn-H/AwesomeWorldModels}{https://github.com/Li-Zn-H/AwesomeWorldModels}.
\end{abstract}
}

\maketitle

\IEEEdisplaynontitleabstractindextext
\IEEEpeerreviewmaketitle

\section{Introduction}\label{sec:introduction}
\IEEEPARstart{E}{mbodied} AI aims to equip agents to perceive complex, multimodal environments, act within them, and anticipate how their actions will alter future world states~\cite{batra2020rearrangement,gupta2024essential}. Central to this capability is the world model, an internal simulator that captures environment dynamics to support both forward and counterfactual rollouts for perception, prediction, and decision making~\cite{liu2025aligning,ding2024understanding}. This survey focuses on world models that yield actionable predictions for embodied agents, distinguishing them from static scene descriptors or purely generative visual models that do not capture controllable dynamics.

Cognitive science suggests humans construct internal models of the world by integrating sensory inputs. These models do not merely predict and simulate future events but also shape perception and guide action~\cite{clark1998being,barsalou1999perceptions,friston2010free}. Motivated by this view, early AI research on world models was rooted in model-based reinforcement learning (RL), where latent state-transition models were used to improve sample efficiency and planning performance~\cite{fung2025embodied}. The seminal work of Ha and Schmidhuber~\cite{ha2018recurrent} crystallized the term world model and inspired the Dreamer series~\cite{hafner2020dream,hafner2021mastering,hafner2025mastering}, highlighting how learned dynamics can drive imagination-based policy optimization. More recently, advances in large-scale generative modeling and multimodal learning have expanded world models beyond their initial focus on policy learning into general-purpose environment simulators capable of high-fidelity future prediction, exemplified by models like Sora~\cite{brooks2024video} and V-JEPA 2~\cite{assran2025v}. This expansion has diversified functional roles, temporal modeling strategies, and spatial representations, while introducing inconsistencies in terminology and taxonomy across sub-communities.

Faithfully capturing environment dynamics requires addressing both the temporal evolution of states and the spatial encoding of scenes~\cite{liu2025aligning}. Long-horizon rollouts are susceptible to error accumulation, which establishes coherence as a central challenge in video prediction and policy imagination~\cite{venkatraman2015improving,asadi2018lipschitz}. Similarly, coarse or 2D-centric layouts provide insufficient geometric detail for handling challenges such as occlusion, object permanence, and geometry-aware planning. In contrast, volumetric or 3D occupancy representations such as neural fields~\cite{agro2024uno} and structured voxel grids~\cite{wei2024occllama} provide geometric structure that better supports forecasting and control. Taken together, these points establish temporal modeling and spatial representation as core design dimensions that fundamentally influence the predictive horizon, physical fidelity, and downstream performance of embodied agents.

Several recent surveys have organized the rapidly growing literature on world models. Overall, these surveys followed two main approaches. The first is a function-oriented perspective. For example, Ding~\etal\cite{ding2024understanding} categorized relevant works based on the two core functions of understanding and prediction, while Zhu~\etal\cite{zhu2024sora} presented a framework based on the core capabilities of world models. The second approach is application-driven, focusing on specific domains such as autonomous driving. Notably, Guan~\etal\cite{guan2024world} and Feng~\etal\cite{feng2025survey} provided overviews of world-model techniques for autonomous driving. Complementing these perspectives, we adopt a three-axis taxonomy around decision coupling, temporal modeling, and spatial representation, and apply it uniformly across robotics, autonomous driving, and general-purpose video world models.

\begin{figure*}[t]
    \centering
    \resizebox{.98\textwidth}{!}{%
        \begin{tikzpicture}[
            box/.style={
                rectangle,  
                draw,                  
                very thick,            
                rounded corners=3pt,   
                align=center,          
                text width=3cm,        
                minimum height=1.5cm,  
                font=\sffamily\bfseries, 
            },
            arrow/.style={
                ->,                      
                thick,                 
                -{Stealth[length=3mm, width=2mm]} 
            },
            node distance=1cm
        ]
        \node[box, fill=intro_fill, draw=intro_border] (intro) {\secref{sec:introduction} Introduction};
        \node[box, fill=bg_fill, draw=bg_border, right=5mm of intro] (background) {\secref{sec:background} Background};
        \node[box, fill=tax_fill, draw=tax_border, right=5mm of background] (taxonomy) {\secref{sec:taxonomy} Taxonomy};
        \node[box, fill=data_fill, draw=data_border, right=5mm of taxonomy] (data) {\secref{sec:data} Data and Metrics};
        \node[box, fill=perf_fill, draw=perf_border, right=5mm of data] (performance) {\secref{sec:performance} Performance};
        \node[box, fill=chall_fill, draw=chall_border, right=5mm of performance] (challenges) {\secref{sec:challenges} Challenges and Trends};
        \node[box, fill=con_fill, draw=con_border, right=5mm of challenges] (conclusion) {\secref{sec:conclusion} Conclusion};

        \draw[arrow] (intro) -- (background);
        \draw[arrow] (background) -- (taxonomy);
        \draw[arrow] (taxonomy) -- (data);
        \draw[arrow] (data) -- (performance);
        \draw[arrow] (performance) -- (challenges);
        \draw[arrow] (challenges) -- (conclusion);

        \begin{scope}[on background layer]
            \node[fit=(intro)(conclusion), inner sep=6pt] (bgfit) {};

            \coordinate (NW) at (bgfit.north west);
            \coordinate (NE) at (bgfit.north east);
            \coordinate (SW) at (bgfit.south west);
            \coordinate (SE) at (bgfit.south east);
            \coordinate (MID) at ($ (NE)!0.5!(SE) + (8mm,0) $);

            \filldraw[
                left color=grad_start,  
                right color=grad_end,   
                opacity=0.97,            
                draw=frame_border_color,          
                line width=0.8pt        
            ](NW) -- (NE) -- (MID) -- (SE) -- (SW) -- cycle;
        \end{scope}
        
        \draw[very thick, draw=gray]($(SW)+(0,-2mm)$) -- ($(SE)+(8mm,-2mm)$);

        \node[anchor=north west, inner sep=0] (concept_fig) at ($(SW)+(5mm,-5mm)$) {
            \includegraphics[height=35mm]{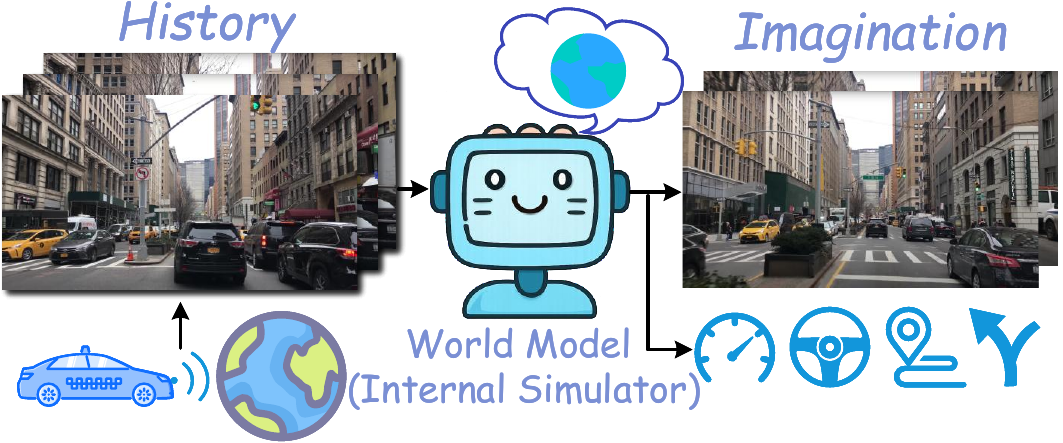}
        };
        \node[
            anchor=north west,
            align=center,
            font=\fontfamily{FiraSans-LF}\selectfont\bfseries\itshape\large, 
            text=wmPurple,
            below=0mm of concept_fig
        ] (concept) {Core Concepts \textup{(\secref{subsec:back_def})}};

        \node[anchor=north west, inner sep=0, right=10mm of concept_fig] (decision_fig) {
            \includegraphics[height=35mm]{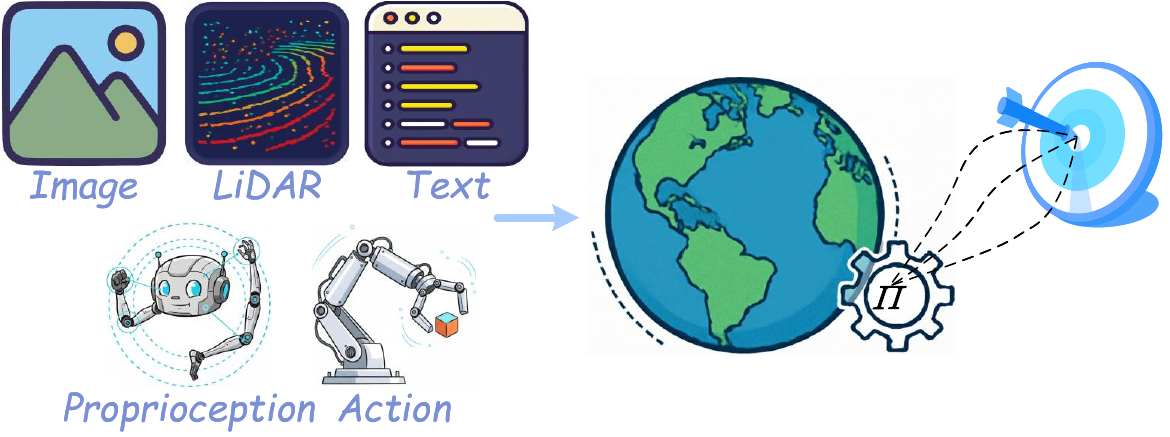} 
        };
        \node[
            anchor=north west, 
            align=center,
            font=\fontfamily{FiraSans-LF}\selectfont\bfseries\itshape\large,
            text=wmPurple,
            below=0mm of decision_fig
        ] (decision) {Decision-Coupled \textup{(\secref{subsec:Decision})}}; 
        
        \node[anchor=north west, inner sep=0, right=8mm of decision_fig] (general_fig) {
            \includegraphics[height=35mm]{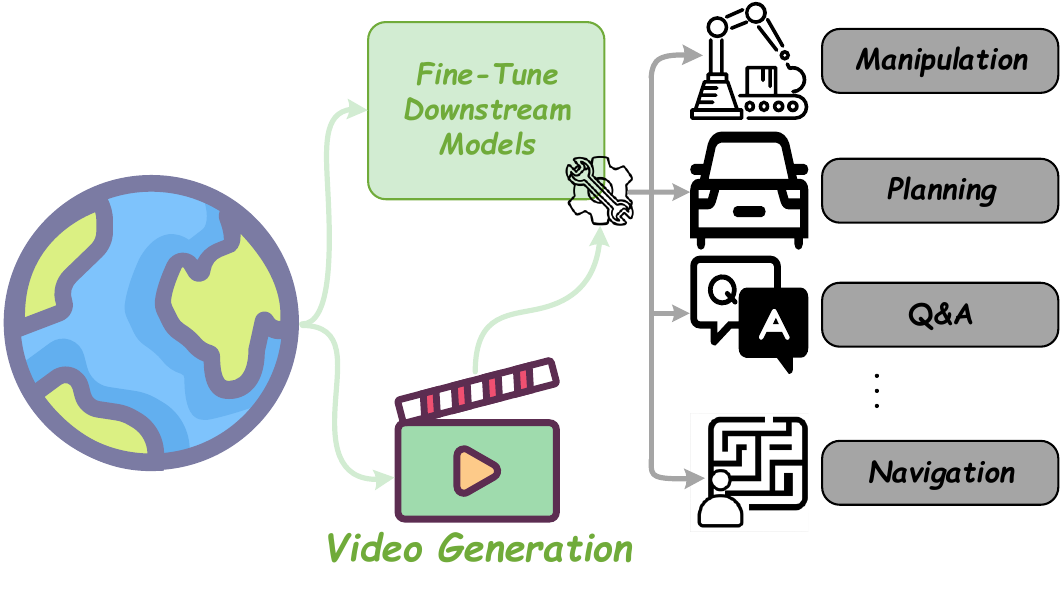} 
        };
        \node[
            anchor=north west, 
            align=center,
            font=\fontfamily{FiraSans-LF}\selectfont\bfseries\itshape\large,
            text=wmPurple,
            below=0mm of general_fig
        ] (general) {General-Purpose \textup{(\secref{subsec:General})}};

        \draw[very thick, draw=gray]($(SW)+(0,-48mm)$) coordinate (line2_left) -- ($(SE)+(8mm,-48mm)$) coordinate (line2_right);

        \node[
            anchor=north west,
            align=center,
            font=\fontfamily{FiraSans-LF}\selectfont\bfseries\itshape\large, 
            text=wmPurple,
            text width=25mm,
        ] (sequential) at ($(line2_left)+(-1mm,-13mm)$) {Sequential \textup{(\secref{subsubsec:Decision_Sequential} \& \secref{subsubsec:General_Sequential})}};
        \node[anchor=north west, inner sep=0, right=-2mm of sequential] (sequential_fig) {
            \includegraphics[height=35mm]{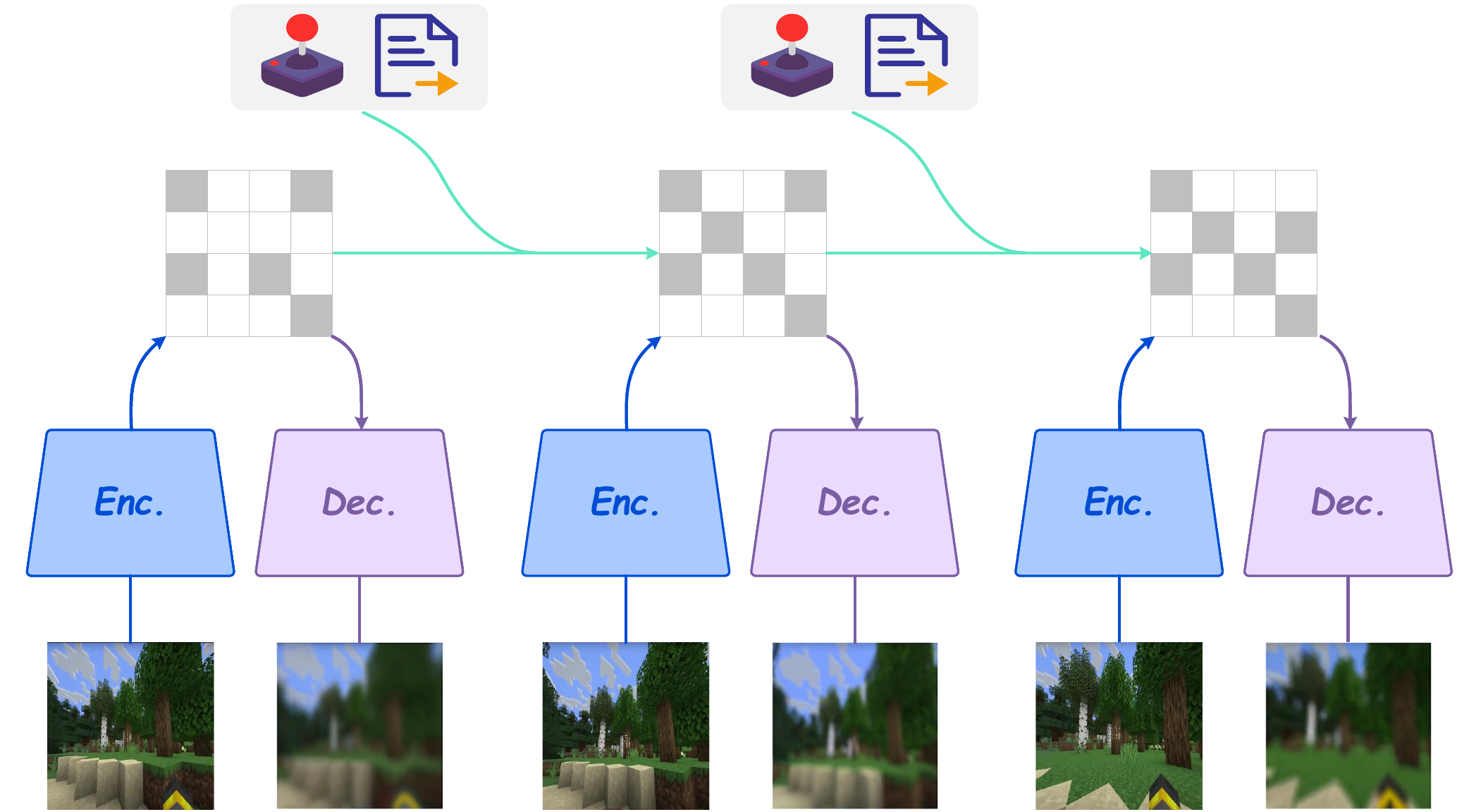} 
        };
        \node[
            anchor=north west, 
            align=center,
            font=\fontfamily{FiraSans-LF}\selectfont\bfseries\itshape\large,
            text=wmPurple,
            below=0mm of sequential_fig
        ] (recurrent) {Recurrent Structure};

        \node[anchor=north west, inner sep=0, right=5mm of sequential_fig] (chunk_fig) {
            \includegraphics[height=35mm]{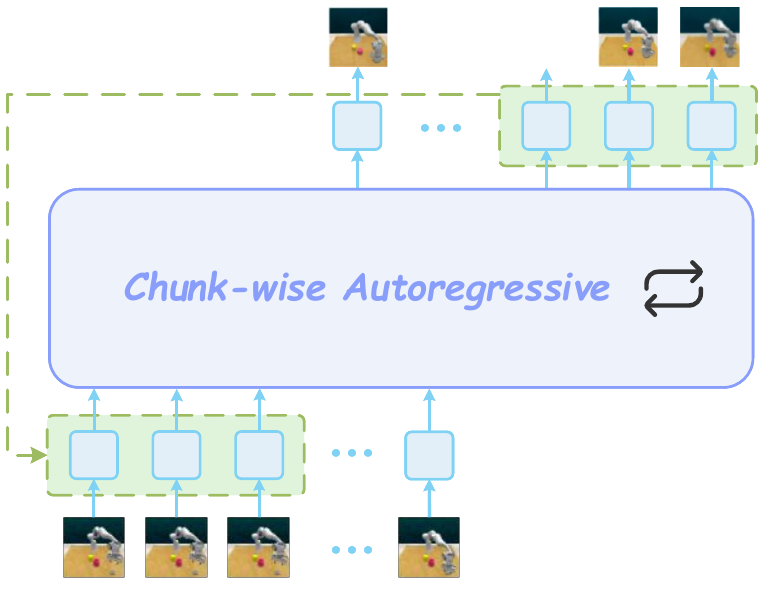} 
        };
        \node[
            anchor=north west, 
            align=center,
            font=\fontfamily{FiraSans-LF}\selectfont\bfseries\itshape\large,
            text=wmPurple,
            below=0mm of chunk_fig
        ] (chunk) {Autoregressive Methods}; 

      \node[
            anchor=north west,
            align=center,
            font=\fontfamily{FiraSans-LF}\selectfont\bfseries\itshape\large, 
            text=wmPurple,
            text width=25mm,
            right=0mm of chunk_fig,
        ] (global) {Global \textup{(\secref{subsubsec:Decision_Global} \& \secref{subsubsec:General_Global})}};

      \node[anchor=north west, inner sep=0, right=0mm of global] (glo_fig) {
            \includegraphics[height=35mm]{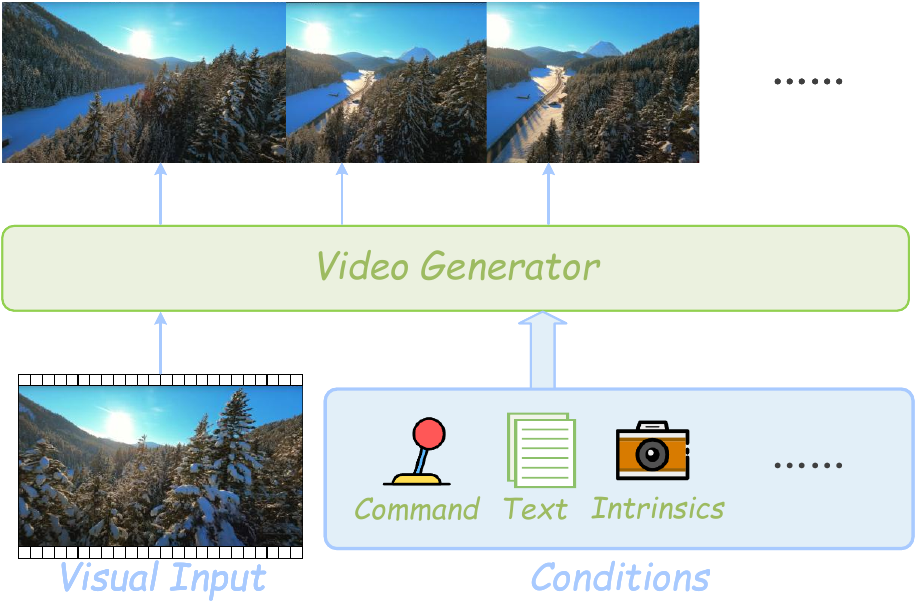} 
        };
      \node[
            anchor=north west, 
            align=center,
            font=\fontfamily{FiraSans-LF}\selectfont\bfseries\itshape\large,
            text=wmPurple,
            below=0mm of glo_fig
        ] (glo) {Global Prediction}; 
        
      \node[anchor=north west, inner sep=0, right=5mm of glo_fig] (mask_fig) {
            \includegraphics[height=35mm]{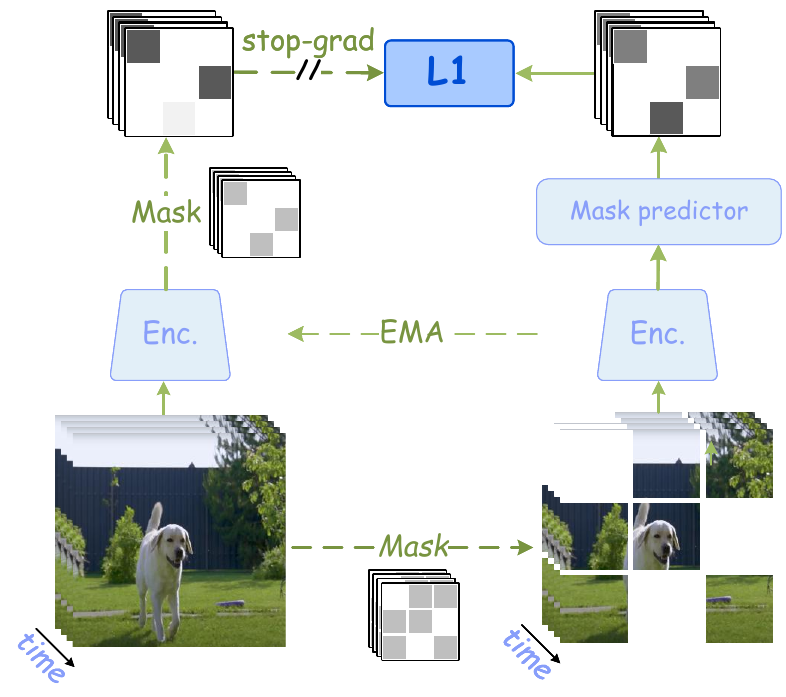} 
        };
      \node[
            anchor=north west, 
            align=center,
            font=\fontfamily{FiraSans-LF}\selectfont\bfseries\itshape\large,
            text=wmPurple,
            below=0mm of mask_fig
        ] (mask) {Masked JEPA}; 

      \draw[very thick, draw=gray]($(line2_left)+(0,-48mm)$) coordinate (line3_left) -- ($(line2_right)+(0mm,-48mm)$) coordinate (line3_right);

      \node[
            anchor=north west,
            align=center,
            font=\fontfamily{FiraSans-LF}\selectfont\bfseries\itshape\large, 
            text=wmPurple,
            text width=25mm,
      ] (Spatial) at ($(line3_left)+(-1mm,-15mm)$) {Spatial\\Representation};

      \node[anchor=north west, inner sep=0, right=-3mm of Spatial] (GLV_fig) {
            \includegraphics[height=35mm]{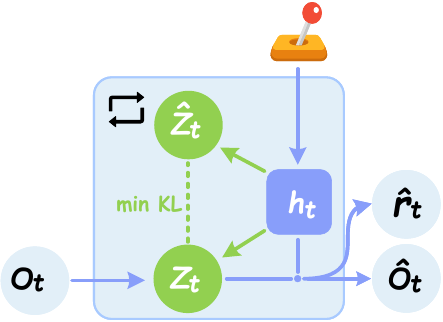} 
      };

      \node[
        anchor=north west, 
        align=center,
        font=\fontfamily{FiraSans-LF}\selectfont\bfseries\itshape\large,
        text=wmPurple,
        below=1mm of GLV_fig
      ] (GLV) {Global Latent Vector}; 

      \node[anchor=north west, inner sep=0, right=0mm of GLV_fig] (TFS_fig) {
            \includegraphics[height=35mm]{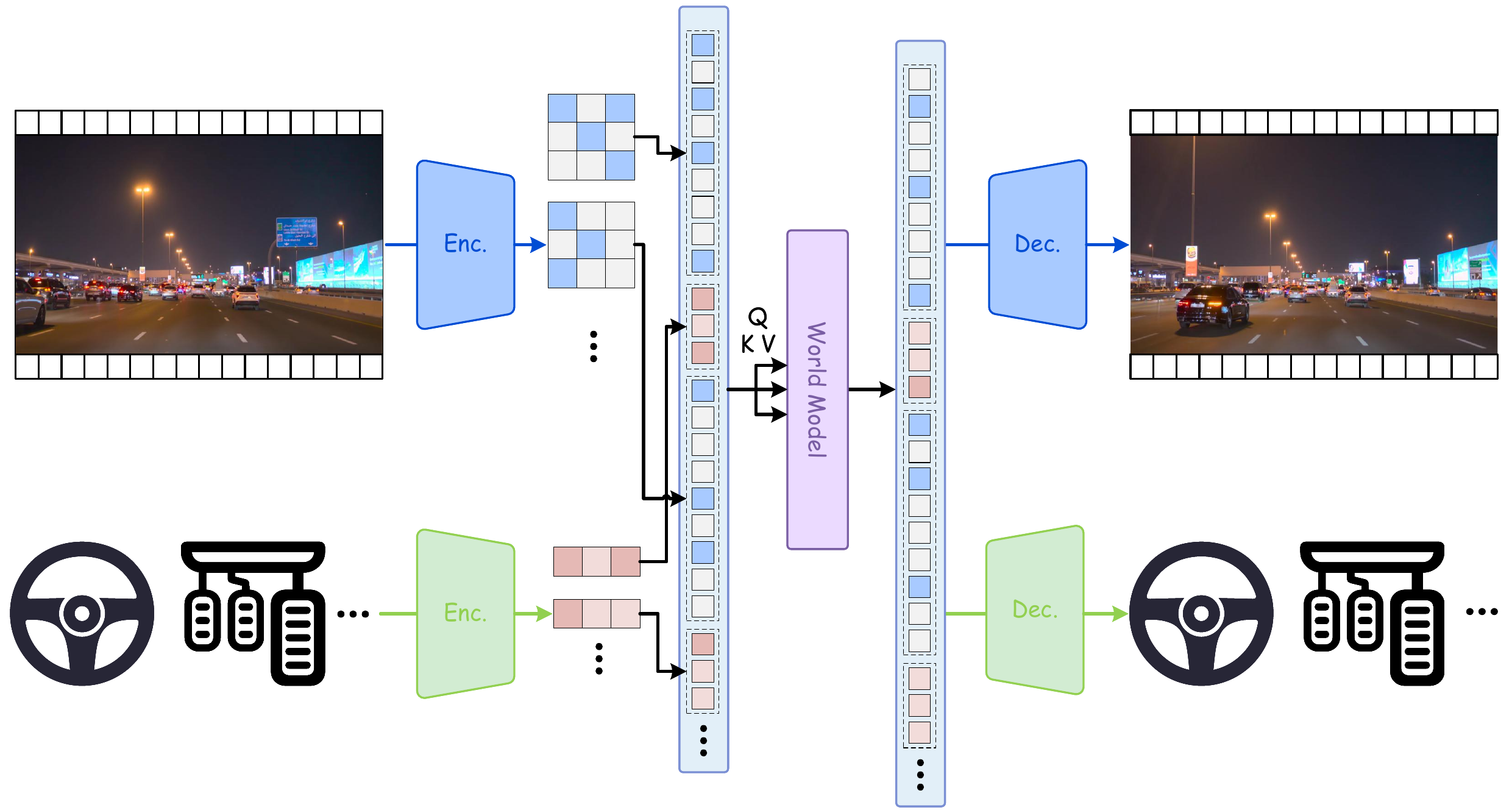} 
      };

      \node[
        anchor=north west, 
        align=center,
        font=\fontfamily{FiraSans-LF}\selectfont\bfseries\itshape\large,
        text=wmPurple,
        below=1mm of TFS_fig
      ] (TFS) {Token Feature Sequence}; 

      \node[anchor=north west, inner sep=0, right=2mm of TFS_fig] (SLG_fig) {
            \includegraphics[height=35mm]{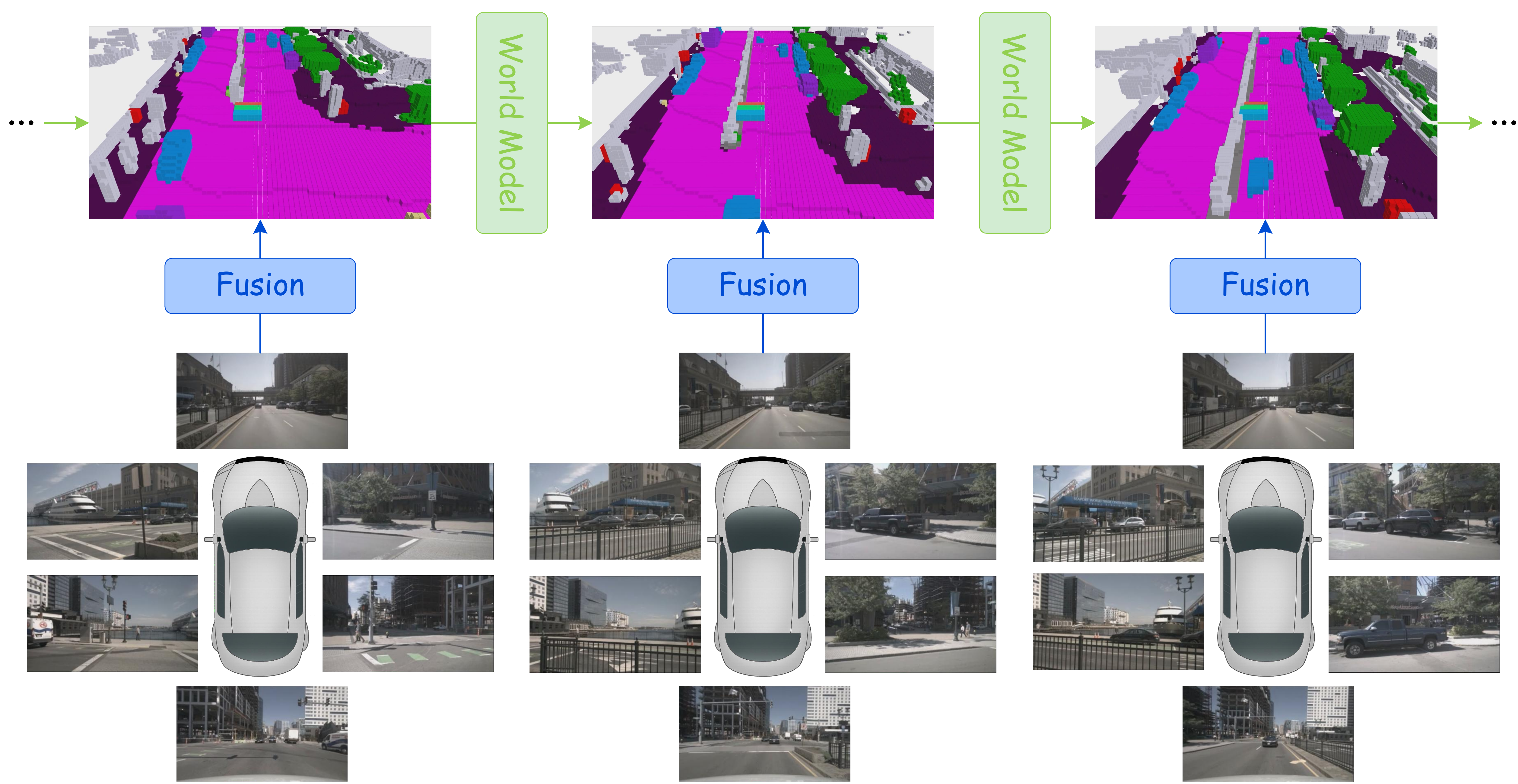} 
      };

      \node[
        anchor=north west, 
        align=center,
        font=\fontfamily{FiraSans-LF}\selectfont\bfseries\itshape\large,
        text=wmPurple,
        below=1mm of SLG_fig
      ] (SLG) {Spatial Latent Grid}; 

      \node[anchor=north west, inner sep=0, right=2mm of SLG_fig] (DRR_fig) {
            \includegraphics[height=35mm]{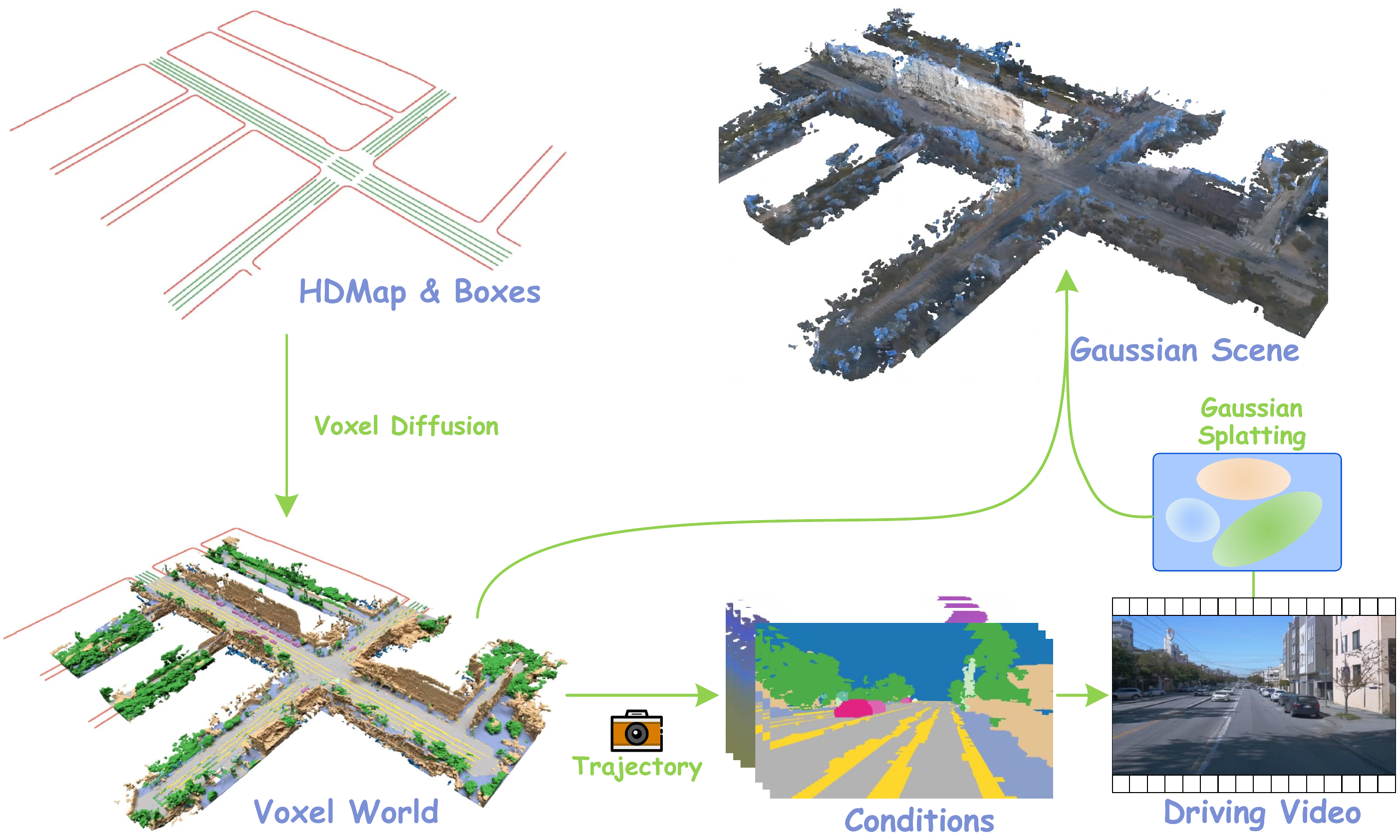} 
      };
  
      \node[
        anchor=north west, 
        align=center,
        font=\fontfamily{FiraSans-LF}\selectfont\bfseries\itshape\large,
        text=wmPurple,
        below=1mm of DRR_fig
      ] (DRR) {Decomposed Rendering Representation}; 

      \end{tikzpicture}
    }
  \caption{Structure of this survey. The figure classifies world models along three axes and illustrates representative methods for each, providing a unified view of the field. Figure design inspired in part by~\cite{assran2025v,hafner2025mastering,lu2025infinicube,chen2024drivinggpt}.}
  \label{fig:structure}
\end{figure*}

To address the lack of a unified taxonomy in the context of embodied AI, this work introduces a framework centered on the three core axes of functionality, temporal modeling, and spatial representation. At the functional level, this framework distinguishes between decision-coupled and general-purpose models. At the temporal level, it differentiates between Sequential Simulation and Inference versus Global Difference Prediction. Finally, at the spatial level, it encompasses a range of representations from latent features to explicit geometry and neural fields. This framework offers a unified structure for organizing existing approaches and integrates standardized datasets and evaluation metrics. This structure facilitates quantitative comparisons and provides a panoramic, actionable knowledge map for future research.

\figref{fig:structure} presents an overview of the structure and taxonomy of this paper. We begin in \secref{sec:background} by outlining the core concepts and theoretical foundations of world models. \secref{sec:taxonomy} introduces our three-axis taxonomy and maps representative methods onto this framework. \secref{sec:data} surveys the datasets and evaluation metrics used for training and assessment. \secref{sec:performance} offers a quantitative comparison of state-of-the-art models. \secref{sec:challenges} discusses open challenges and promising research directions, and \secref{sec:conclusion} concludes the survey.

\section{Background}\label{sec:background}
\subsection{Core Concepts}\label{subsec:back_def}
As discussed in \secref{sec:introduction}, a world model can be viewed as an internal simulator that implements a \textbf{compression--prediction--control} pipeline: high-dimensional observations are first \textbf{compressed} into a latent state, the dynamics model \textbf{predicts} how this state will evolve under
actions, and a policy or planner \textbf{controls} the agent by reasoning over imagined latent rollouts. Operationally, this functionality decomposes into three tightly coupled aspects: 
\begin{itemize}[leftmargin=*, nosep]
    \item \textbf{Simulation \& Planning}, which uses learned dynamics to generate plausible scenarios, allowing agents to assess potential actions through imagination without real-world interaction.
    \item \textbf{Temporal Evolution}, which learns how the encoded state evolves, enabling temporally consistent rollouts.
    \item \textbf{Spatial Representation}, which encodes scene geometry at an appropriate fidelity, using formats such as latent tokens or neural fields to provide context for control.
\end{itemize}

Together, these three pillars instantiate the compression--prediction--control view and jointly determine how a world model navigates the fundamental trade-off between computational efficiency, long-horizon coherence, and visual fidelity. They provide the conceptual foundation for the taxonomy introduced in \secref{sec:taxonomy} and are formalized in the mathematical framework that follows.

\subsection{Mathematical Formulation of World Models}\label{subsec:back_math}
We formalize the environment interaction as a POMDP~\cite{smallwood1973optimal}. For notational consistency, we define a null initial action \(a_0\) at \(t=0\), which allows the dynamics to be written uniformly. At each step \(t \ge 1\), the agent receives an observation \(o_t\) and takes an action \(a_t\), while the true state \(s_t\) remains unobserved. To handle this partial observability, world models infer a learned latent state \(z_t\) using a one-step filtering posterior, where the previous latent state $z_{t-1}$ is assumed to summarize the relevant history. Finally, $z_t$ is used to reconstruct $o_t$:
\begin{equation}
\label{eq:world_model}
\begin{array}{ll}
\text{Dynamics Prior:} & p_\theta(z_t \mid z_{t-1}, a_{t-1}) \\
\text{Filtered Posterior:} & q_\phi(z_t \mid z_{t-1}, a_{t-1}, o_t) \\
\text{Reconstruction:} & p_\theta(o_t \mid z_t)
\end{array}.
\end{equation}
Consistent with the Markovian structure, the joint distribution over observations and latent states factorizes as:
\begin{equation}
\label{eq:gen}
p_\theta(o_{1:T}, z_{0:T} \!\mid\! a_{0:T-1}) \!=\! p_\theta(z_0)\!\prod_{t=1}^{T}\! p_\theta(z_t \!\mid\! z_{t-1}, a_{t-1})p_\theta(o_t \!\mid\! z_t).
\end{equation}
To infer the latent states, we must approximate the intractable true posterior $p_\theta(z_{0:T}\!\mid\! o_{1:T},a_{0:T-1})$ with a time-factorized variational distribution:
\begin{equation}
\label{eq:post}
q_\phi(z_{0:T}\!\mid\! o_{1:T}, a_{0:T-1}) = q_\phi(z_0\!\mid\! o_1)\!\prod_{t=1}^{T}\! q_\phi(z_t\!\mid\! z_{t-1}, a_{t-1}, o_t),
\end{equation}
which indeed reduces to the action-free case when ignoring $a$ inputs. Directly maximizing the log-likelihood $\log p_\theta(o_{1:T}\!\mid\! a_{0:T-1})$ is intractable. Instead, we optimize an ELBO using the approximate posterior $q_\phi$, which provides a tractable objective for learning the model parameters:  
\begin{equation}
\begin{aligned}
\label{eq:elbo1}
\log p_\theta(&o_{1:T}\!\mid\! a_{0:T-1}) = \!\log\!\int p_\theta(o_{1:T},z_{0:T}\!\mid\! a_{0:T-1})\,dz_{0:T} \\
&\ge  \mathbb{E}_{q_\phi}\Big[\log \frac{p_\theta(o_{1:T},z_{0:T}\mid a_{0:T-1})}{\,q_\phi(z_{0:T}\mid o_{1:T},a_{0:T-1})}\Big] =: \mathcal{L}(\theta,\phi).
\end{aligned}
\end{equation}
Under the assumption of Markov factorization for both $p_\theta$ and $q_\phi$, this ELBO decomposes into a reconstruction objective and a KL regularization term:  
\begin{equation}
\label{eq:elbo_short}
\begin{aligned}
\mathcal{L}(\theta,&\phi) = \sum_{t=1}^{T}\mathbb{E}_{q_\phi(z_t)}\!\big[\log p_\theta(o_t\mid z_t)\big]\\
&
- D_{\mathrm{KL}}\!\big(q_\phi(z_{0:T}\mid o_{1:T},a_{0:T-1})\,\|\,p_\theta(z_{0:T}\mid a_{0:T-1})\big).
\end{aligned}
\end{equation}
Modern world models thus adopt a reconstruction-regularization training paradigm: the likelihood term $\log p_\theta(o_t\!\mid\! z_t)$ encourages faithful observation prediction, and KL regularization terms align the filtered posterior $q_\phi(z_t\!\mid\! z_{t-1},a_{t-1},o_t)$ with the dynamics prior $p_\theta(z_t\!\mid\! z_{t-1},a_{t-1})$. Such world models can be instantiated with recurrent models~\cite{sun2024learning,zhai2025recurrent,aljalbout2025accelerating}, Transformer-based architectures~\cite{chen2022transdreamer,robine2023transformerbased,zhang2023storm}, or diffusion-based decoders~\cite{zhu2025unified,qi2025strengthening,guo2025flowdreamer,huang2025enerverse,alhaija2025cosmos}. In all cases, the learned latent trajectory $z_{1:T}$ serves as a compact, predictive memory to support downstream policy optimization, model-predictive control, and counterfactual reasoning in embodied AI.

\section{Taxonomy}\label{sec:taxonomy}
We categorize world models along three core dimensions, which provide the foundation for the subsequent analysis in this survey.

The first dimension, decision coupling, distinguishes between \textbf{Decision-Coupled} and \textbf{General-Purpose} world models. Decision-Coupled models are task-specific, learning dynamics tightly aligned with a particular control objective, which improves sample efficiency and closed-loop performance but weakens generalization beyond the training distribution. In contrast, General-Purpose models are task-agnostic simulators that prioritize broad coverage of visual and physical phenomena, easing transfer across tasks and domains, yet their pretraining objective often misaligns with downstream control, leading to suboptimal or brittle behavior unless explicitly mitigated.

The second dimension, temporal reasoning, delineates two distinct paradigms of prediction. \textbf{Sequential Simulation and Inference} models dynamics autoregressively, unfolding future states one step at a time. This offers fine-grained control and naturally supports closed-loop planning, but suffers from error accumulation over long horizons and a computational cost that scales linearly with the rollout length. In contrast, \textbf{Global Difference Prediction} directly estimates future states in parallel, amortizing computation across time and reducing wall-clock latency for long-horizon imagination, but its global updates weaken closed-loop interactivity and tend to smooth over fine-grained local dynamics, making it less suited to tasks that require frequent feedback and precise step-wise control.

The third dimension, spatial representation, comprises four primary strategies used in current research to model spatial states. They lie at different points along the coherence-fidelity-efficiency trade-off:
\begin{enumerate}
    \item \textbf{Global Latent Vector} encodes world states into a compact vector, enabling efficient real-time rollouts but sacrificing fine-grained spatial detail.
    \item \textbf{Token Feature Sequence} discretizes world states into token sequences to model token-token dependencies via attention, enabling fine-grained multimodal modeling but typically requiring large data, large models, and high inference cost.
    \item \textbf{Spatial Latent Grid} injects spatial priors by encoding scenes into BEV or voxel grids, preserving local topology and supporting multi-view fusion and map-like planning, but incurring large memory cost, limited resolution, and weaker adaptability to unstructured environments.
    \item \textbf{Decomposed Rendering Representation} decomposes scenes into renderable primitives (\eg, 3DGS~\cite{kerbl20233d}, NeRF~\cite{mildenhall2020nerf}) and reconstructs observations via differentiable rendering, achieving geometry-consistent, high-fidelity and object-level controllable views, but with expensive training and limited ability to handle rapid dynamics or topology changes.
\end{enumerate}
The following tables apply this taxonomy to classify representative works. \tabref{tab:Summary_1} reviews approaches in robotics, while \tabref{tab:Summary_2} focuses on autonomous driving. Together, they provide a roadmap for the detailed analyses in the subsequent sections.

\begin{table*}[!t]
\centering
\caption{\textbf{A summary of representative world models in robotics and general-purpose domains.}}
\label{tab:Summary_1}
\renewcommand{\arraystretch}{1.15}
\rowcolors{5}{gray!10}{white}
\resizebox{\linewidth}{!}{%
\begin{tabular}{lccccccccccccccccccccc} %
\toprule[1.5pt]
\multirow{2}{*}[-5ex]{\textbf{Paper}} &
\multirow{2}{*}[-5ex]{\textbf{Publication}} &
\multirow{2}{*}[-5ex]{\textbf{Taxonomy}\textsuperscript{\protect\hyperlink{note:Taxonomy_1}{1}}} &
\multirow{2}{*}[-5ex]{\textbf{Characteristics}\textsuperscript{\protect\hyperlink{note:Characteristics_1}{2}}} &
\multicolumn{11}{c}{\textbf{Datasets Platform}} & %
\multicolumn{6}{c}{\textbf{Modality}} & %
\multirow{2}{*}[-5ex]{\textbf{Reality}\textsuperscript{\protect\hyperlink{note:Reality_1}{4}}}  \\
\cmidrule(lr){5-15} %
\cmidrule(lr){16-21} %
 & & & & 
\rot{DMC} & \rot{Atari}  & \rot{RLBench} & \rot{SSv2} & \rot{OXE} & \rot{Meta-world} & \rot{Franka} & \rot{RT-1} & \rot{LIBERO} & \rot{Other(s)} & \rot{\textbf{Total}\textsuperscript{\protect\hyperlink{note:Total_1}{3}}} &
\rot{RGB} & \rot{Action} & \rot{Proprio.} & \rot{Depth}  & \rot{Language} & \rot{Other(s)} & 
 \\
\specialrule{0.5pt}{2pt}{0pt}

PlaNet\cite{hafner2019learning} & ICML'19  & Dec/Seq/GLV & RSSM &
\cmark &  & & & & & & & & & \textbf{1} & 
\cmark & \cmark & & & & & \\

Dreamer\cite{hafner2020dream} & ICLR'20  & Dec/Seq/GLV & RSSM & 
\cmark & \cmark  & & & &  & & & & \cmark & \textbf{3} & 
\cmark& \cmark & & & & & \\

GLAMOR\cite{paster2021planning} & ICLR'21  & Dec/Seq/GLV & IDM & 
\cmark & \cmark  & & & & &  &  & & & \textbf{2} & 
\cmark & \cmark & &  & & & \\

DreamerV2\cite{hafner2021mastering} & ICLR'21  & Dec/Seq/GLV & RSSM & 
\cmark & \cmark  & & &  & &   & &  & & \textbf{2} & 
\cmark & \cmark & &  & & & \\

TransDreamer\cite{chen2022transdreamer} & arXiv'22   & Dec/Seq/GLV & TSSM & 
\cmark & \cmark  & & & & & & & & \cmark & \textbf{4} & 
\cmark & \cmark  & &  & & & \\

Iso-Dream\cite{pan2022iso} & NeurIPS'22   & Dec/Seq/GLV & IDM &
\cmark &   & & &  & &  &  & & \cmark & \textbf{4} & 
\cmark & \cmark & &  & & & \\

MWM\cite{seo2023masked} & CoRL'22  & Dec/Seq/TFS & RSSM & 
\cmark &   & \cmark  & &   & \cmark & &  & &   & \textbf{3} & 
\cmark & \cmark & &  & & & \\
Inner Monologue\cite{huang2023inner} & CoRL'22  & Dec/Seq/TFS & CoT & 
 &   &   & &  & & &  &  & \cmark & \textbf{3} & 
\cmark & \cmark & &  & \cmark & & \cmark  \\
DayDreamer\cite{wu2023daydreamer} & CoRL'22  & Dec/Seq/GLV & RSSM & 
 &  & & & &  & &  &  & \cmark & \textbf{4} & 
\cmark & \cmark & \cmark & \cmark  & & & \cmark  \\

TWM\cite{robine2023transformerbased} & ICLR'23   & Dec/Seq/TFS & Transformer & 
 & \cmark  & & & &  &  & &  &  & \textbf{1} & 
\cmark & \cmark &  &   & & & \\
IRIS\cite{micheli2023transformers} & ICLR'23 & Dec/Seq/TFS & Transformer & 
 & \cmark  & & & &  &  & &  &  & \textbf{1} & 
\cmark & \cmark &  &   & & & \\

WorldDreamer\cite{wang2024worlddreamer} & arXiv'24 & Gen/Glo/TFS & Transformer & 
 &  &  &  & &  &  &  &  & \cmark & \underline{\textbf{4}} & 
\cmark & \cmark &  &  & \cmark  & & 
 \\

Statler\cite{yoneda2024statler} & ICRA'24  & Dec/Seq/TFS & LLM & 
 &   & & & &  & &  &  & \cmark & \textbf{2} & 
\cmark & \cmark &  &  \cmark & \cmark & & 
\cmark  \\

Pandora\cite{xiang2024pandora} & arXiv'24  & Gen/Seq/TFS & Video Diffusion & 
 &  &  & \cmark & &  &  &  &  & \cmark & \underline{\textbf{2}} & 
\cmark & &  &  & \cmark  & & 
 \\

DWL\cite{gu2024advancing} & RSS'24  & Dec/Seq/GLV & MLP  & 
 &  & & &  & & &  &  & \cmark & \textbf{4} & 
 & \cmark & \cmark &   & & \cmark & 
\cmark \\

RoboDreamer\cite{zhou2024robodreamer} & ICML'24  & Dec/Glo/TFS & IDM & 
 &  & \cmark &  & &  &  &  \cmark &  & & \textbf{2} & 
\cmark&\cmark & &  & \cmark & \cmark & 
  \\
Genie\cite{bruce2024genie}   & ICML'24 & Gen/Seq/TFS & Transformer & 
 &  &  &  & &  &  &  \cmark &  & \cmark & \underline{\textbf{3}} & 
\cmark& \cmark & &  & & & 
 \\

V-JEPA\cite{bardes2024revisiting}  & TMLR'24  & Gen/Glo/TFS & JEPA & 
 &   &  & \cmark &  & &  &  &  & \cmark & \underline{\textbf{6}} & 
\cmark &  &   &   &  &   & 
  \\

PreLAR\cite{zhang2024prelar} & ECCV'24  & Dec/Seq/GLV & RSSM &
 &    & \cmark & \cmark &  & \cmark &  &  &  &  & \textbf{3} &
\cmark & \cmark &  &  &  &  &
 \\
ManiGaussian\cite{lu2024manigaussian} & ECCV'24  & Dec/Seq/DRR & 3DGS & 
 &  & \cmark &  & &  &  &  &  &  & \textbf{1} & 
\cmark & \cmark & \cmark & \cmark &  \cmark & & 
 \\

ECoT\cite{zawalski2025robotic} &  CoRL'24 & Dec/Glo/TFS & CoT & 
 &  &  &  & &   &  &  &  & \cmark & \textbf{3} & 
\cmark& \cmark &  &  & \cmark & & 
\cmark  \\

VidMan\cite{wen2024vidman} & NeurIPS'24 & Dec/Glo/TFS & IDM & 
 &  & \cmark &  &\cmark &  &  &  &  & \cmark & \textbf{4} & 
\cmark& \cmark& \cmark &  & \cmark & & 
  \\
iVideoGPT\cite{wu2024ivideogpt} & NeurIPS'24 & Gen/Seq/TFS & Transformer & 
 &  &  & \cmark &\cmark & \cmark &  &  &  & \cmark & \textbf{6} & 
\cmark & \cmark &  &  &  & & 
  \\

EnerVerse\cite{huang2025enerverse} & arXiv'25 & Dec/Seq/SLG & Video Diffusion & 
 &  & & &  & &  & \cmark & \cmark  & \cmark & \underline{\textbf{4}} & 
\cmark & \cmark &  &  & \cmark &  & 
\cmark  \\

GLAM\cite{he2025glam} & AAAI'25  & Dec/Seq/GLV & Mamba & 
 & \cmark  &  &  & &  & &  &  &  & \textbf{1} & 
\cmark & \cmark  & &   & & & 
 \\

NavCoT\cite{lin2025navcot} & TPAMI'25  & Dec/Seq/TFS & CoT & 
 &   &  &  & &  &  & &  & \cmark & \textbf{4} & 
\cmark & \cmark  & &   & \cmark & & 
  \\

DreamerV3\cite{hafner2025mastering} & Nature'25  & Dec/Seq/GLV & RSSM & 
\cmark & \cmark & & &  & & &  &  & \cmark & \textbf{8} & 
\cmark & \cmark & \cmark &  & & & 
 \\

MineWorld\cite{guo2025mineworld} & arXiv'25   & Dec/Seq/TFS & Transformer & 
 &  & & &  & & &  &  & \cmark & \textbf{1} & 
\cmark & \cmark &  &  &  & & 
  \\

DreMa\cite{barcellona2025dream} & ICLR'25  & Dec/Seq/DRR & 3DGS & 
 &   & \cmark & &  & & \cmark &  &  &  & \textbf{2} & 
\cmark & \cmark &  & \cmark & \cmark & \cmark & 
\cmark \\

S2-SSM\cite{petri2025learning} & arXiv'25 & Gen/Seq/TFS & Mamba & 
 &  &  &  &  & &  &  &  & \cmark & \textbf{1} & 
\cmark &  &   &  & & \cmark & 
 \\

RLVR-World\cite{wu2025rlvr} & arXiv'25 & Gen/Seq/TFS & RLVR & 
 &  &  &  &  & & \cmark &  &  & \cmark & \textbf{3} & 
\cmark & \cmark &   &  &\cmark & & 
  \\

StateSpaceDiffuser\cite{savov2025statespacediffuser} & arXiv'25 & Gen/Seq/TFS & Mamba & 
 &  &  &  &  & &  &  &  & \cmark & \textbf{2} & 
\cmark &\cmark  &   &  & & & 
  \\

DeepVerse\cite{chen2025deepverse} & arXiv'25 & Gen/Seq/TFS & DiT & 
 &  &  & &  & &  & &  &  \cmark & \underline{\textbf{1}} & 
\cmark &  & &  \cmark &  \cmark & \cmark & 
 \\

ORV\cite{yang2025orv}  & arXiv'25 & Gen/Glo/SLG & DiT & 
 &  &  &  &  & &  & \cmark &  & \cmark & \textbf{4} & 
\cmark & \cmark &   & \cmark  & \cmark &  \cmark & 
 \\

V-JEPA 2\cite{assran2025v}  & arXiv'25 & Gen/Glo/TFS & JEPA & 
 &  &  & \cmark &  & & \cmark &  &  & \cmark & \underline{\textbf{15}} & 
\cmark & \cmark & \cmark  &   & \cmark &   & 
\cmark \\

NWM\cite{bar2025navigation} & CVPR'25 & Dec/Seq/TFS & DiT & 
 &    & & &  & & &  &  & \cmark & \textbf{6} & 
\cmark & \cmark &  &  & &  & 
  \\

WorldVLA\cite{cen2025worldvla} & arXiv'25  & Dec/Seq/TFS & Transformer & 
 &   & & &  & & &  & \cmark &  & \textbf{1} & 
\cmark & \cmark &  &  & \cmark & & 
 \\
World4Omni\cite{chen2025world4omni} & arXiv'25 & Gen/Seq/TFS & VLM & 
 &  & \cmark &  &  & &  &  &  & \cmark & \textbf{2} & 
\cmark & \cmark &   &   & \cmark & \cmark  & 
\cmark \\

Dyn-O\cite{wang2025dyn} & arXiv'25  & Dec/Seq/TFS & Mamba & 
 &   & & &  & &  & &  & \cmark & \textbf{1} & 
\cmark & \cmark &  &  & &  & 
  \\

DINO-WM\cite{zhou2025dino} & ICML'25 & Dec/Seq/SLG & Transformer & 
\cmark &   & & &  & & &  &  & \cmark & \textbf{3} & 
\cmark & \cmark & \cmark &  &  & & 
  \\
EVA\cite{chi2025empowering} & ICML'25 & Gen/Seq/TFS & RoG & 
&   & & & \cmark & & & \cmark  &  & \cmark & \underline{\textbf{4}} & 
\cmark & \cmark & &  &  \cmark & & 
 \\
AdaWorld\cite{gao2025adaworld} & ICML'25 & Gen/Seq/TFS & Video Diffusion & 
&  & & \cmark& \cmark &  & &   & \cmark & \cmark & \underline{\textbf{6}} & 
\cmark & \cmark & &  &   & & 
 \\

MindJourney\cite{yang2025mindjourney}  & arXiv'25 & Gen/Seq/SLG & VLM & 
 &  &  &  &  & &  &  &  & \cmark & \underline{\textbf{2}} & 
\cmark & \cmark &   &   & \cmark &   & 
 \\

GAF\cite{chai2025gaf} & arXiv'25  & Dec/Seq/DRR & 4DGS & 
 &  & \cmark & &  & &  & &  &  & \textbf{1} & 
\cmark & \cmark & \cmark&  &  & \cmark & 
 \\

Yume\cite{mao2025yume} & arXiv'25 & Gen/Seq/TFS & DiT & 
 &  &  &  &  & &  &  &  & \cmark & \textbf{1} & 
\cmark & \cmark  &   &  &\cmark & & 
 \\

villa-X\cite{chen2025villa} & arXiv'25 & Dec/Glo/TFS & IDM & 
 &  &  & \cmark & \cmark & &  & \cmark & \cmark & \cmark & \underline{\textbf{5}} & 
\cmark & \cmark & \cmark  &  & \cmark & & 
\cmark  \\

AETHER\cite{team2025aether}  & ICCV'25  & Gen/Glo/SLG & DiT & 
 &   &  &  &  & &  &  &  & \cmark & \underline{\textbf{6}} & 
\cmark & \cmark &   & \cmark  &  & \cmark  & 
 \\
 TesserAct\cite{zhen2025tesseract}  & ICCV'25 & Dec/Glo/SLG & IDM & 
 &   & \cmark & \cmark &  & &  & \cmark &  & \cmark & \textbf{4} & 
\cmark &  &   & \cmark & \cmark & \cmark & 
  \\

MineDreamer\cite{zhou2025minedreamer}& IROS'25  & Dec/Seq/TFS & CoI & 
 &    & & &  & & &   &  & \cmark & \underline{\textbf{3}} & 
\cmark &\cmark  &  &  & \cmark & & 
 \\
ManiGaussian++\cite{yu2025manigaussian++} & IROS'25 & Dec/Seq/DRR & 3DGS & 
 &   & \cmark & &  &  & &  &  & \cmark & \textbf{2} & 
\cmark & \cmark & \cmark & \cmark & \cmark & & 
\cmark  \\

LeWorldModel\cite{maes2026leworldmodel} & arXiv'26 & Dec/Seq/GLV & JEPA & 
\cmark &  &  &  &  &  &  &  &  & \cmark & \textbf{4} & 
\cmark & \cmark &  &  &  &  & 
 \\

DreamZero\cite{ye2026world} & arXiv'26 & Dec/Seq/TFS & Video Diffusion & 
 &  &  &  &  &  & \cmark &  &  & \cmark & \underline{\textbf{4}} & 
\cmark & \cmark & \cmark &  & \cmark &  & 
\cmark  \\

PointWorld\cite{huang2026pointworld} & CVPR'26 & Dec/Seq/SLG & Transformer & 
 &  &  &  &  &  & \cmark &  &  & \cmark & \underline{\textbf{2}} & 
\cmark & \cmark & \cmark & \cmark &  & \cmark & 
\cmark  \\

Cosmos Policy\cite{kim2026cosmos} & arXiv'26 & Dec/Glo/TFS & Video Diffusion & 
 &  &  &  &  &  & \cmark &  & \cmark & \cmark & \underline{\textbf{3+}} & 
\cmark & \cmark & \cmark &  & \cmark & \cmark & 
\cmark  \\

V-JEPA 2.1\cite{mur2026v} & arXiv'26 & Gen/Glo/TFS & JEPA & 
 &  &  & \cmark &  &  & \cmark &  &  & \cmark & \underline{\textbf{15+}} & 
\cmark & \cmark & \cmark & \cmark & \cmark & \cmark & 
\cmark  \\

Cosmos 3\cite{agarwal2026cosmos} & arXiv'26 & Gen/Glo/TFS & Transformer & 
 &  &  &  &  &  & \cmark & \cmark & \cmark & \cmark & \underline{\textbf{12+}} & 
\cmark & \cmark & \cmark & \cmark & \cmark & \cmark & 
\cmark  \\

\specialrule{1.5pt}{0pt}{0pt}
\end{tabular}}
\begin{tablenotes}[flushleft]\fontsize{6pt}{7pt}\selectfont
    \item \hypertarget{note:Taxonomy_1}{}\textsuperscript{1} \textbf{Taxonomy}: Abbreviations for the taxonomy categories defined in \secref{sec:taxonomy}.
    \item \hypertarget{note:Characteristics_1}{}\textsuperscript{2} \textbf{Characteristics}: Representative backbone or core technical approach.
    \item \hypertarget{note:Total_1}{}\textsuperscript{3} \textbf{Total}: Number of data platforms used. Underlined entries denote newly proposed or aggregated datasets.
    \item \hypertarget{note:Reality_1}{}\textsuperscript{4} \textbf{Reality}: The check mark (\cmark) indicates validation on a physical robot.
\end{tablenotes}
\vspace{-1.5mm}
\end{table*}

\begin{table*}[!t]
\centering
\caption{\textbf{A summary of representative world models for the autonomous driving domain.}}
\label{tab:Summary_2}
\renewcommand{\arraystretch}{1.1}
\rowcolors{5}{gray!10}{white}
\resizebox{\linewidth}{!}{%
\begin{tabular}{lccccccccccccccccccc} %
\toprule
\multirow{2}{*}[-5ex]{\textbf{Paper}} &
\multirow{2}{*}[-5ex]{\textbf{Publication}} &
\multirow{2}{*}[-5ex]{\textbf{Taxonomy}\textsuperscript{\protect\hyperlink{note:Taxonomy_2}{1}}} &
\multirow{2}{*}[-5ex]{\textbf{Characteristics}\textsuperscript{\protect\hyperlink{note:Characteristics_2}{2}}} &
\multicolumn{8}{c}{\textbf{Datasets Platform}} & %
\multicolumn{7}{c}{\textbf{Input Modality}} \\
\cmidrule(lr){5-12} %
\cmidrule(lr){13-20} %
 & &  & &
\rot{CARLA} & \rot{nuScenes} & \rot{nuPlan} & \rot{Waymo} & \rot{Occ3D} & \rot{OpenDV} & \rot{Other(s)} & \rot{\textbf{Total}\textsuperscript{\protect\hyperlink{note:Total_2}{3}}} &
\rot{RGB} & \rot{Motion}   & \rot{Map} & \rot{LiDAR} &\rot{Bound box} & \rot{Language} & \rot{Occpancy} & \rot{Other(s)}  \\
\specialrule{0.5pt}{3pt}{0pt}

MILE\cite{hu2022model} & NeurIPS'22 & Dec/Seq/GLV & RSSM &
\cmark & & & & & & & \textbf{1} & 
\cmark & \cmark   & \cmark & & & & & \\
Copilot4D\cite{zhang2024copilot4d} & ICLR'24   & Gen/Seq/SLG & Video Diffusion &
 & \cmark &  & & &  & \cmark & \textbf{3} & 
 &  \cmark     &   & \cmark &  & & &  \\

SEM2\cite{gao2024enhance} & TITS'24  & Dec/Seq/GLV & RSSM &
\cmark & & & & & & & \textbf{1} & 
\cmark & \cmark    & \cmark & \cmark & & & &  \\

MagicDrive3D\cite{gao2024magicdrive3d} & arXiv'24  & Gen/Glo/DRR & 3DGS &
 & \cmark &  & &  & &  & \textbf{1} & 
\cmark & \cmark     & \cmark &  & \cmark & \cmark &  & \cmark  \\

OccSora\cite{wang2024occsora} & arXiv'24   & Gen/Glo/SLG & Diffusion &
 & \cmark &  & & \cmark &  &  & \textbf{2} & 
 &    \cmark  &  &  &  &  & \cmark &    \\

Delphi\cite{ma2024unleashing} & arXiv'24  & Gen/Seq/SLG & Video Diffusion &
 & \cmark & & &   & &  & \textbf{1} & 
 \cmark &      & \cmark &  & \cmark & \cmark &  &    \\

DriveWorld\cite{min2024driveworld} & CVPR'24 & Dec/Seq/SLG & RSSM &
 & \cmark & & & & & \cmark & \textbf{2} & 
\cmark &  \cmark  &  &  &  & \cmark & \cmark & \\
Drive-WM\cite{wang2024driving} & CVPR'24  & Dec/Glo/SLG & Video Diffusion &
 & \cmark & & & & &  & \textbf{1} & 
\cmark &   \cmark   & \cmark  &  & \cmark & \cmark & & \cmark \\
ViDAR\cite{Yang2024visual} & CVPR'24 & Gen/Seq/SLG & Transformer &
 & \cmark & & & & &  & \textbf{1} & 
\cmark & \cmark      &   & \cmark &  &  & &  \\
GenAD\cite{yang2024generalized} & CVPR'24  & Gen/Seq/TFS & Video Diffusion &
 & \cmark & \cmark &\cmark & & \cmark & \cmark & \underline{\textbf{4}} & 
\cmark & \cmark     &   &  &  & \cmark & &\\

OccLLaMA\cite{wei2024occllama} & arXiv'24  & Dec/Seq/SLG & Transformer &
 & \cmark & & & \cmark  & & \cmark & \underline{\textbf{3}} & 
\cmark &   \cmark  &  &  &  & \cmark & \cmark &\\

DriveDreamer\cite{wang2024drivedreamer} & ECCV'24  & Dec/Seq/SLG & GRU &
 & \cmark & & & & & & \textbf{1} & 
\cmark &    \cmark  & \cmark &  & \cmark & \cmark & & \\
GenAD\cite{zheng2024genad} & ECCV'24   & Dec/Seq/SLG & GRU &
 & \cmark & & & & & & \textbf{1} & 
\cmark &      &  &  &  &  & & \\
OccWorld\cite{zheng2024occworld} & ECCV'24   & Dec/Seq/SLG & Transformer &
 & \cmark & & & \cmark & & & \textbf{2} & 
\cmark & \cmark     &  & \cmark &  &  & \cmark &  \\

DOME\cite{gu2024dome} & arXiv'24   & Gen/Seq/SLG & DiT &
 & \cmark & & & \cmark  & &  & \textbf{2} & 
\cmark &   \cmark  &  &  &  &  & \cmark &  \\

TOKEN\cite{tian2025tokenize} & CoRL'24  & Dec/Glo/TFS & Transformer &
 & \cmark &  & & & & \cmark & \textbf{2} & 
\cmark & \cmark     & \cmark &  &  & \cmark & &    \\

Vista\cite{gao2024vista} &  NeurIPS'24 & Gen/Seq/SLG & Video Diffusion &
 & \cmark &  & \cmark & & \cmark& \cmark & \textbf{4} & 
\cmark &  \cmark      &   &  &  & \cmark & &   \\

DriveDreamer-2\cite{zhao2025drivedreamer} & AAAI'25  & Gen/Glo/SLG & Video Diffusion &
 & \cmark &  & &  & &  & \textbf{1} & 
\cmark &    \cmark  & \cmark &  & \cmark & \cmark &  &  \\

DTT\cite{xu2025delta} & arXiv'25  & Dec/Seq/DRR & Transformer &
 & \cmark &  & & \cmark & &  & \textbf{2} & 
\cmark &    \cmark  &  &  &  & \cmark & \cmark &  \\

DynamicCity\cite{bian2025dynamiccity} & ICLR'25  & Gen/Glo/SLG & DiT &
\cmark & \cmark &  & \cmark & \cmark &  &  & \textbf{4} & 
\cmark &    \cmark  &  &  &  & \cmark & \cmark &  \\

LidarDM\cite{zyrianov2025lidardm}& ICRA'25  & Gen/Seq/SLG & Diffusion &
 &  &  & \cmark &  & & \cmark & \textbf{3} & 
 &    \cmark  & \cmark &  &  &  & &   \\

FutureSightDrive\cite{zeng2025futuresightdrive} & arXiv'25 & Dec/Seq/TFS & CoT(VLM) &
 & \cmark &  & &  & & \cmark & \textbf{3} & 
\cmark &  \cmark    &  &  &  & \cmark &  & \\

GEM\cite{hassan2025gem} & CVPR'25  & Gen/Seq/SLG & Video Diffusion &
 & \cmark &  &  & & \cmark &  & \underline{\textbf{1}} & 
\cmark  &    \cmark  &  &  &  &  & & \cmark   \\
GaussianWorld\cite{zuo2025gaussianworld} & CVPR'25   & Gen/Seq/DRR & Transformer &
 & \cmark &  &  & & & \cmark & \textbf{2} & 
\cmark  & \cmark     &  &  &  &  & &    \\
MaskGWM\cite{ni2025maskgwm} & CVPR'25  & Gen/Glo/TFS & DiT &
 & \cmark &  & \cmark & & \cmark &  & \textbf{3} & 
\cmark  &     \cmark  &  &  &  & \cmark & &   \\
DriveDreamer4D\cite{zhao2025drivedreamer4d} & CVPR'25  & Gen/Glo/DRR & 4DGS &
 & \cmark   & \cmark &  & & \cmark & \textbf{3} & 
\cmark &   & \cmark  & \cmark & \cmark & \cmark & \cmark &  & \cmark  \\
ReconDreamer\cite{ni2025recondreamer} & CVPR'25   & Gen/Glo/DRR & 3DGS &
 & \cmark &  & \cmark &  & & \cmark & \textbf{3} & 
\cmark &    \cmark  & \cmark & \cmark & \cmark &  &  &    \\

WoTE\cite{li2025end} & ICCV'25  & Dec/Seq/SLG & Transformer &
\cmark &  & \cmark & & & &  & \textbf{2} & 
\cmark &    \cmark  &  & \cmark &  &  & & \\
HERMES\cite{zhou2025hermes} & ICCV'25   & Gen/Glo/SLG & LLM &
 & \cmark &  & &  &  & \cmark & \textbf{4} & 
\cmark &    \cmark  &  &  &  & \cmark &  &   \\
InfiniCube\cite{lu2025infinicube} & ICCV'25  & Gen/Seq/DRR & 3DGS &
 &  &  & \cmark & & &  & \textbf{1} & 
\cmark  &    \cmark   & \cmark &  & \cmark & \cmark & & \cmark   \\

DriVerse\cite{li2025driverse}& ACMMM'25  & Gen/Seq/TFS & DiT &
 & \cmark &  & \cmark & & &  & \textbf{2} & 
\cmark &    \cmark  &  &  &  & \cmark & &  \cmark \\

DINO-Foresight\cite{karypidis2026dino} & NeurIPS'25 & Gen/Glo/TFS & Transformer &
 & \cmark &  &  &  &  & \cmark & \textbf{2} & 
\cmark &  &  &  &  &  &  & \cmark \\

\specialrule{1.5pt}{0pt}{0pt}
\end{tabular}}
\begin{tablenotes}[flushleft]\fontsize{6pt}{7pt}\selectfont
    \item \hypertarget{note:Taxonomy_2}{}\textsuperscript{1} \textbf{Taxonomy}: Abbreviations for the taxonomy categories defined in \secref{sec:taxonomy}.
    \item \hypertarget{note:Characteristics_2}{}\textsuperscript{2} \textbf{Characteristics}: Representative backbone or core technical approach.
    \item \hypertarget{note:Total_2}{}\textsuperscript{3} \textbf{Total}: Number of data platforms used. Underlined entries denote newly proposed or aggregated datasets.
\end{tablenotes}
\end{table*}

\subsection{Decision-Coupled World Models}
\label{subsec:Decision}
\subsubsection{Sequential Simulation and Inference}\label{subsubsec:Decision_Sequential}
\textbf{Global Latent Vector.}\quad Early decision-coupled world models combined sequential inference with global latent states. These approaches primarily use Recurrent Neural Networks (RNNs) for efficient real-time and long-horizon prediction.

Ha and Schmidhuber~\cite{ha2018recurrent} introduced an early world model that encodes observations into a latent space and employs an RNN to model dynamics for policy optimization. Building on this, PlaNet~\cite{hafner2019learning} introduced the Recurrent State-Space Model (RSSM), which fuses deterministic memory with stochastic components to enable robust long-horizon imagination. The successor models Dreamer, DreamerV2, and DreamerV3~\cite{hafner2020dream, hafner2021mastering, hafner2025mastering} further advanced this formulation, inspiring a broad line of subsequent research.

Building on RSSM, several variants modify or eliminate the decoder to better capture dynamics. For example, Dreaming~\cite{okada2021dreaming} uses contrastive learning and linear methods to mitigate state shifts, whereas DreamerPro~\cite{deng2022dreamerpro} replaces the decoder with prototypes to suppress visual distractions. To further enhance robustness, HRSSM~\cite{sun2024learning} was proposed, featuring a dual-branch architecture that aligns latent observations and shares information without reconstruction. Beyond architectural refinements, DisWM~\cite{wang2025disentangled} disentangles semantic knowledge from video content, distilling it into a world model that enables cross-domain generalization. LeWorldModel~\cite{maes2026leworldmodel} extends this compact-latent line beyond RSSM-style objectives by jointly training a pixel encoder and action-conditioned Transformer predictor with a next-embedding loss and SIGReg anti-collapse regularization for reward-free latent planning.

A unifying theme across recent RSSM extensions is transferability, which reflects generalization across modalities, tasks, and embodiments for robust real-world robotics. At the representation level, PreLAR~\cite{zhang2024prelar} learns implicit action abstractions to bridge video-pretrained representations and control fine-tuning. Similarly, Wang~\etal\cite{wang2025latent} used optical flow as an embodiment-agnostic action representation to refine behavior-cloned policies, facilitating transfer across diverse embodiments. SENSEI~\cite{sancaktar2025sensei} distilled a Vision-Language Model (VLM) to derive semantic rewards and employed an RSSM that learns to predict and propagate these rewards internally. Under limited supervision, SR-AIF~\cite{nguyen2025sr} exploits prior preference learning and self-revision to enable adaptive learning in sparse-reward, continuous-control settings. To mitigate the Sim-to-Real (S2R) gap, ReDRAW~\cite{lanier2025adapting} is pretrained in simulation and adapted to the real environment using a limited amount of reward-free data, applying residual corrections to the latent dynamics. To handle mismatches, AdaWM~\cite{wang2025adawm} identifies discrepancies between the learned dynamics and the planner and selectively fine-tunes critical components. Other methods like WMP~\cite{lai2025world} address S2R transfer for challenging tasks, and DayDreamer~\cite{wu2023daydreamer} demonstrated sample-efficient deployment on physical robots. To broaden transfer,  FOUNDER~\cite{wang2025founder} grounds representations from foundation models in the world-model state space, using temporal-distance prediction to handle flexible goals, and LUMOS~\cite{nematollahi2025lumos} introduced a language-conditioned imitation framework that operates on-policy in latent space with intrinsic rewards, enabling zero-shot transfer to real-world robotics.

RSSM-based models have also been developed for autonomous driving. MILE~\cite{hu2022model} leverages offline expert data to enable imagined future states for planning. SEM2~\cite{gao2024enhance} integrates semantic filtering with multi-source sampling to extract driving-relevant features and balance data distribution. Popov~\etal\cite{popov2025mitigating} addressed covariate shift through a latent generative world model that realigns policies with expert states. For safety, VL-SAFE~\cite{qu2025vl} supervises world models using safety scores derived from a VLM to generate safe trajectories. Finally, CALL~\cite{wang2025ego} extended the RSSM framework to Multi-Agent RL by introducing ego-centric information sharing to enhance planning capabilities.

In contrast to RSSM, TransDreamer~\cite{chen2022transdreamer} introduced a Transformer State-Space Model (TSSM) that replaces the recurrent core in Dreamer, thereby substantially enhancing its capacity to capture long-horizon dependencies. The complementary OSVI-WM~\cite{goswami2025osvi} employs a causal Transformer for one-shot Imitation Learning (IL), autoregressively predicting the future latent trajectory and decoding it into physical waypoints for robot control. 

Some approaches continue to employ RNNs to capture temporal dependencies. On the modeling side, RWM~\cite{li2025robotic} introduced a dual-autoregressive, domain-agnostic neural simulator for long-horizon prediction. X-MOBILITY~\cite{liu2025x}, in contrast, disentangles modeling from policy learning, using multi-head decoders for large-scale pretraining followed by supervised fine-tuning to derive strong policies. For humanoid locomotion, DWL~\cite{gu2024advancing} and WMR~\cite{sun2025learning} adopted end-to-end (E2E) frameworks. These frameworks reconstruct states from partial observations using either denoising or a gradient-blocked state estimator, which enables zero-shot transfer across complex real-world terrains.

Recently, State Space Models (SSMs), exemplified by Mamba, have emerged as alternatives to RNNs and Transformers, combining linear-time complexity with long-horizon modeling capacity. Building on this, GLAM~\cite{he2025glam} improves both fidelity and efficiency via a Mamba-based parallel framework that integrates global and local modules to capture contextual and fine-grained dynamics.

Beyond forward temporal modeling, Inverse Dynamics Modeling (IDM) is a key paradigm in world model construction. An IDM infers the actions required to transition between an initial and a target state. Agrawal~\etal\cite{agrawal2016learning} integrated a forward model with an IDM for multi-step prediction, establishing the basis for subsequent research. More recent work includes GLAMOR~\cite{paster2021planning}, which trains an object-conditioned IDM to predict the actions necessary to reach a specified target. In Dreamer-style agents, Iso-Dream\cite{pan2022iso} leverages IDM to decompose the world model into controllable and uncontrollable components, using the rollout of uncontrollable states to guide policy learning.

\para{Token Feature Sequence.} The Token Feature Sequence paradigm centers on modeling dependencies among discrete tokens. This representation supports causal inference, multimodal integration, and the reuse of Large Language Model (LLM).

Recent RSSM-centric studies have begun to exploit token-level dependencies to strengthen representation learning and temporal reasoning. For example, MWM~\cite{seo2023masked} decouples visual tokens from RSSM-based dynamics via a masked autoencoder, improving both performance and data efficiency. NavMorph~\cite{yao2025navmorph} introduced a self-evolving RSSM with a contextual evolution memory for online adaptation. For temporal abstraction, WISTER~\cite{burchi2025learning} employed action-conditioned contrastive predictive coding to train a TSSM that captures high-level temporal features. Similarly, TWM~\cite{robine2023transformerbased} used a Transformer to align multimodal tokens with historical states during training, while relying on a lightweight policy at inference. To handle long-horizon tasks, some approaches integrate LLMs with RSSMs to decompose objectives into subtasks. EvoAgent~\cite{feng2025evoagent}, for example, uses an LLM to guide low-level actions and regularizes RSSM updates. RoboHorizon~\cite{chen2025robohorizon}, in contrast, enhances task recognition with dense rewards and leverages key task segments via a masked autoencoder.

In autonomous driving, token-based sequence representations are increasingly adopted to model cross-modal interactions and spatiotemporal structures. DrivingWorld~\cite{hu2024drivingworld} pairs next-state prediction for temporal dynamics with next-token prediction for spatial structure. For multimodal control, Doe-1~\cite{zheng2024doe} formulates closed-loop driving as autoregressive prediction over perception-description-action tokens, which unifies perception, prediction, and planning, and DrivingGPT~\cite{chen2024drivinggpt} interleaves vision and action tokens and casts world modeling and trajectory planning as next-token prediction. To enhance diversity and safety, LatentDriver~\cite{xiao2025learning} models future actions as a mixture distribution and actuates the world model with planner-sampled intermediate actions. At the same time, Vasudevan~\etal\cite{vasudevan2025planning} proposed an adaptive model that predicts surrounding agents for safe navigation.

The token-based paradigm also extends to broader robotics. Within RL, IRIS~\cite{micheli2023transformers} and TWM~\cite{dedieu2025improving} leverage discrete tokens to enable data-efficient policy learning via imagined or hybrid rollouts. DyWA~\cite{lyu2025dywa} improves action learning by conditioning on trajectory dynamics and jointly predicting future states with single-view point-cloud and proprioceptive modalities. EgoAgent~\cite{chen2025egoagent} interleaves state-action sequence modeling within a Transformer, enabling unified perception, prediction, and action inference. Tokenized representations unify multimodal inputs, including vision, language, and action (VLA), enabling generalist agents with cross-domain adaptability, as shown in WorldVLA~\cite{cen2025worldvla}. DreamZero~\cite{ye2026world} further instantiates this paradigm as a World Action Model built on a pretrained video diffusion backbone, autoregressively predicting future video and actions from language, visual history, and proprioception to support real-time closed-loop robot control. Recent studies encode environmental states as discrete symbolic tokens and condition next-token prediction on action, as demonstrated by DCWM~\cite{scannell2025discrete} and TrajWorld~\cite{yin2025trajectory}.

Recent studies have strengthened the link between tokenized representations and planning, particularly through object-centric approaches. These models, such as CarFormer~\cite{hamdan2024carformer}, the work of Jeong~\etal\cite{jeong2025object}, and Dyn-O~\cite{wang2025dyn}, represent scenes as a collection of slots. CarFormer autoregressively models the relationships between these slots in BEV. Jeong~\etal added language-guided manipulation, while Dyn-O used a Mamba with dropout scheduling for temporal modeling and to disentangle static from dynamic elements. 
$\Delta$-IRIS~\cite{micheli2024efficient} introduced a hybrid Transformer that integrates tokens with stochastic $\Delta$-tokens to capture dynamics. $\text{D}^2$PO~\cite{wang2025world} employed preference learning to jointly optimize state prediction and action selection, enhancing the model's understanding of underlying dynamics. For efficiency, MineWorld~\cite{guo2025mineworld} accelerated token generation by predicting sequences in parallel and introduced an IDM as a controllability metric. Meanwhile, PIVOT-R~\cite{zhang2024pivot} and ReOI~\cite{chen2025reimagination} incorporated VLMs into control. PIVOT-R parses instructions to produce waypoint-based plans that an action module decodes into low-level controls, whereas ReOI detects implausible prediction elements, reimagines the distractors, and reintegrates the corrected content.

Based on tokenization, some studies employ autoregressive diffusion to achieve stable generation and long-horizon planning. Epona~\cite{zhang2025epona} decouples spatiotemporal modeling, handled by a Transformer, from long-horizon multimodal generation, realized through trajectory and vision Diffusion Transformers (DiTs). Goff~\etal\cite{goff2025learning} used a DiT to instantiate the state transition, which enables on-policy training and multi-second closed-loop rollouts.  SceneDiffuser++\cite{tan2025scenediffuser++} pushes this further to city-scale traffic simulation, applying multi-tensor diffusion over agents and traffic lights to produce stable closed-loop rollouts. For navigation, NWM~\cite{bar2025navigation} introduced an efficient conditional DiT to simulate visual trajectories for zero-shot planning.  

Another emerging direction is to inject explicit reasoning into the world model using LLMs and Chain-of-Thought (CoT). NavCoT~\cite{lin2025navcot} decomposes navigation into imagination, filtering, and prediction, enabling parameter-efficient in-domain training, and ECoT~\cite{zawalski2025robotic} leverages a pipeline of foundation models to generate reasoning labels for training a VLA policy. Variants like MineDreamer~\cite{zhou2025minedreamer} introduced Chain-of-Imagination (CoI), where a multimodal LLM imagines future observations to steer diffusion and guide actions, and FSDrive~\cite{zeng2025futuresightdrive} generates physics-constrained future scenes and treats them as CoT supervision, enabling VLMs to function as IDMs for planning.

Other approaches directly couple  LLMs with world models to operationalize planning and data generation. Dyna-Think~\cite{yu2025dyna} fuses reasoning and acting via a distilled LLM, and RIG~\cite{zhao2025rig} unifies reasoning and imagination end-to-end generalist policy. In terms of explicit dynamics and long-horizon, Gkountouras~\etal\cite{gkountouras2025language} trained a causal world-model simulator that grounds an LLM nvironment causal reasoning and planning skills, Statler~\cite{yoneda2024statler} enables LLMs to keep a structured world-state, using a reader for planning and a writer for updating, and Inner Monologue~\cite{huang2023inner} incorporates a closed-loop feedback into LLMs, enabling agents to reason and deliberate more akin to human thinking. Finally, WoMAP~\cite{yin2025womap} synthesized 3DGS scenes and trained a world model that refines VLM instructions for precise execution.

\para{Spatial Latent Grid.} By encoding features on geometry-aligned grids or incorporating explicit spatial priors, this paradigm preserves locality, enables efficient convolutional or attention-based updates and streaming rollouts. 

In autonomous driving, many studies couple RNN-based dynamics with spatial grids to guide planning. For instance,  DriveDreamer\cite{wang2024drivedreamer} and GenAD\cite{zheng2024genad} adopt GRU-based dynamics over grid or instance-centric tokens to predict motion and decode trajectories. In contrast, DriveWorld~\cite{min2024driveworld} and Raw2Drive~\cite{yang2025raw2drive} instantiate RSSM dynamics on BEV tokens. DriveWorld conditions on tokens and  actions for joint prediction, while Raw2Drive adopts a dual-stream design for spatiotemporal learning.

Numerous studies focus on autoregressively predicting future 3D occupancy representations to enable motion planning for autonomous driving. One strand discretizes scenes into occupancy tokens for sequential prediction, exemplified by OccWorld~\cite{zheng2024occworld} and RenderWorld~\cite{yan2025renderworld}. Another strand directly forecasts volumetric features or embeddings, as in Drive-OccWorld~\cite{yang2025driving} and PreWorld~\cite{li2025semi}. Self-supervised variants predict future representations from current cues. For example, LAW~\cite{li2025enhancing} conditions on current representations and trajectories, SSR~\cite{li2025navigation} compresses scenes into sparse BEV tokens for future BEV features, and NeMo~\cite{huang2024neural} voxelizes multi-frame images and predicts occupancy to support imitation-based planning. Building on these representations, FASTopoWM~\cite{yang2025fastopowm} employs a unified decoder to align fast and slow systems from vehicle poses, enabling lane-topology reasoning, and WoTE~\cite{li2025end} simulates candidate trajectories in BEV and selects among them with a reward model. Extending the paradigm, OccLLaMA~\cite{wei2024occllama} unifies occupancy, actions, and text within a single token vocabulary and employs a LLaMA for next token forecasting, planning, and question answering.

Beyond autonomous driving, similar formulations have been extended to broader domains of robotics. WMNav~\cite{nie2025wmnav} leverages a VLM to maintain a curiosity-driven value map and adopts phased decision making to enable zero-shot, object-driven navigation. RoboOccWorld~\cite{zhang2025occupancy} targets indoor robotics by predicting fine-grained 3D occupancy with a pose-conditioned autoregressive Transformer, thereby supporting exploration and decision making. To achieve high-fidelity dynamics, EnerVerse~\cite{huang2025enerverse} applies chunk-wise autoregressive video diffusion with a sparse memory mechanism to produce 4D latent dynamics and integrates 4DGS to mitigate the S2R gap in robotic execution. For manipulation, ParticleFormer~\cite{huang2025particleformer} forecasts future point clouds with a Transformer-based particletized dynamics model, enabling robust handling of multi-object and multi-material interactions. Scaling this point-flow view, PointWorld~\cite{huang2026pointworld} unifies scene states and robot actions as 3D point flows, using a point-cloud Transformer backbone to predict action-conditioned full-scene displacements for MPC-based real-world manipulation from RGB-D observations. At the representation level, DINO-WM~\cite{zhou2025dino} learns dynamics in the DINOv2 feature space and predicts future states to support zero-shot planning.

\para{Decomposed Rendering Representation.} This paradigm represents scenes with explicit renderable primitives such as NeRFs and 3DGS, updating them to simulate dynamics and render future observations. It provides view-consistent forecasts, object-level compositionality, and seamless integration with physics priors and digital twins, thereby supporting long-horizon rollouts.

Building on 3DGS, GAF~\cite{chai2025gaf} augments splats with learnable motion attributes to forecast future states and refines initial actions with diffusion. ManiGaussian~\cite{lu2024manigaussian} predicts per-point variations to generate future Gaussian scenes for manipulation under current states and actions, and ManiGaussian++~\cite{yu2025manigaussian++} adds a hierarchical leader-follower design with task-oriented splats to model primitive deformations for multi-body and bimanual skills. Within simulation and digital twin coupling, DreMa~\cite{barcellona2025dream} integrates GS with a physics simulator to build twins for data synthesis in imitation learning, Abou-Chakra~\etal\cite{abou2025physically} introduced a dual Gaussian-Particle representation where Gaussian points are attached to particles driven by visual loss forces, DexSim2Real$^2$~\cite{jiang2025dexsim2real} builds twins of articulated objects with generative models and uses sampling based planning for precise manipulation, PIN-WM~\cite{li2025pin} combines 3DGS with differentiable physics to estimate physical parameters from limited observations and generates digital cousins for zero-shot S2R policy learning, and PWTF~\cite{ning2025prompting} constructs an interactive twin that simulates candidate action outcomes and uses VLMs for evaluation and selection. At the representation level, DTT~\cite{xu2025delta} adopts a triplane representation with multiscale Transformers to autoregressively capture incremental changes, forming a 4D world model for prediction and planning.

\subsubsection{Global Difference Prediction}\label{subsubsec:Decision_Global} 
\textbf{Token Feature Sequence.}\quad  The compact Global Latent Vector representation discards fine-grained spatiotemporal detail and is therefore rarely used for global prediction. In contrast, Token Feature Sequences predict the future sequence in parallel, reducing error accumulation while enabling multimodal diversity.

On the representation side, TOKEN~\cite{tian2025tokenize} tokenizes scenes into object-level tokens, aligning world representations with reasoning and leveraging LLMs to predict full future trajectories for long-tail scenarios. GeoDrive~\cite{chen2025geodrive} extracts 3D representations, renders trajectory-conditioned views, and edits the position of vehicles to guide DiT in producing editable generations. For control, FLARE~\cite{zheng2025flare} aligns diffusion policies with latent future representations, avoiding pixel-space video generation and learning effectively from action-free videos. In a related vein, LaDi-WM~\cite{huang2025ladi} predicts future states through interactive diffusion in a latent space aligned with visual foundation models, integrating geometric and semantic features while iteratively refining the diffusion policy to improve performance and generalization. Cosmos Policy~\cite{kim2026cosmos} adapts Cosmos-Predict2 into a unified visuomotor policy by injecting proprioception, action chunks, future observations, and values as latent frames, allowing one video diffusion model to serve as policy, world model, and value function for model-based planning. villa-X~\cite{chen2025villa} and VidMan~\cite{wen2024vidman} both couple diffusion-based models with IDM for control. villa-X infers latent actions, aligns them with ego-centric forward dynamics, and maps them via joint diffusion, while VidMan adapts a pretrained video-diffusion model into an IDM using a self-attention adapter for accurate action prediction.

\para{Spatial Latent Grid.} Spatial grid models parallel-forecast BEV or voxel maps from ego-stabilized views, preserving locality and uncertainty while producing planner-ready maps for fast control.

Diffusion-based world models are commonly used for parallel generation. EmbodiedDreamer~\cite{wang2025embodiedreamer} couples differentiable physics with video diffusion to render photorealistic and physically consistent futures. TesserAct~\cite{zhen2025tesseract} reconstructs 4D spatiotemporal consistent scenes by jointly generating RGB, depth, and normal videos for IDM-based action learning. DFIT-OccWorld~\cite{zhang2024efficient} reformulates prediction as decoupled voxel warping and adopts image-assisted single-stage training for reliable and efficient dynamic scene modeling. For instruction-conditioned control, RoboDreamer~\cite{zhou2024robodreamer} decomposes instructions into low-level primitives that steer video diffusion, synthesizing novel compositional scenes beyond the training distribution while grounding execution via an IDM, and ManipDreamer~\cite{li2025manipdreamer} extends this design with an action-tree prior together with depth and semantic guidance to improve instruction following and temporal consistency.

On the planning side, 3DFlowAction~\cite{zhi20253dflowaction} employs a pretrained 3D optical flow world model to treat future motion as a unified action cue, enabling label-free and cross-robot manipulation through closed-loop optimization. Imagine-2-Drive~\cite{garg2025imagine} integrates video diffusion with a multimodal diffusion policy to accelerate policy learning. Drive-WM~\cite{wang2024driving} uses multi-view diffusion with image-based rewards to select safer trajectories, while World4Drive~\cite{zheng2025world4drive} leverages vision-based priors to construct intent-aware world models that support self-supervised multi-intent imagination. COMBO~\cite{zhang2025combo} composes multi-agent actions with diffusion, leverages VLMs to infer purposes, and integrates tree search for online cooperative planning.

\subsection{General-Purpose World Models}\label{subsec:General}
\subsubsection{Sequential Simulation and Inference}\label{subsubsec:General_Sequential}
\textbf{Token Feature Sequence.}\quad General-purpose models pretrain task-agnostic dynamics to capture environmental physics and generate future scenes, prioritizing transferability over specific tasks.

Some general world models increasingly pretrain on unlabeled video and use tokenized latent space for robust forecasting and generation. iVideoGPT~\cite{wu2024ivideogpt} was pretrained on large-scale interaction videos for action-free forecasting and later adapted to downstream control. Genie~\cite{bruce2024genie} learned discrete latent actions and spatiotemporal tokens, enabling user-controllable interactive environments via autoregressive dynamics. RoboScape~\cite{shang2025roboscape} jointly learned video generation with temporal depth and keypoint dynamics to improve physical realism. PACT~\cite{bonatti2023pact} tokenized multimodal perception and action and trained a causal Transformer to obtain a unified representation for diverse tasks, and DINO-world~\cite{baldassarre2025back} learns generalizable dynamics by predicting the temporal evolution of DINOv2 features from large-scale unlabeled video corpora. Building on language priors, EVA~\cite{chi2025empowering} introduced a Reflection-of-Generation (RoG) policy that used a VLM for iterative self-correction, strengthening long-horizon anticipation. In the same vein, Owl-1~\cite{huang2024owl} employs a VLM to forecast world dynamics conditioned on current states and generated fragments, explicitly guiding the subsequent fragments and enabling coherent long-horizon video synthesis, while World4Omni~\cite{chen2025world4omni} employs a reflective world model, where a VLM refines subgoal images from an image generator, and integrates them with pretrained modules for zero-shot robotic manipulation.

Recent work adapts video diffusion models into controllable world models that autoregressively imagine future scenes. AdaWorld~\cite{gao2025adaworld} introduced an action-aware pretraining scheme that extracted self-supervised latent actions between adjacent frames to condition diffusion, enabling efficient transfer with minimal interaction. Vid2World~\cite{huang2025vid2world} adapts pretrained video diffusion models into autoregressive interactive world models via causalization and a causal action-guidance mechanism. GenAD~\cite{yang2024generalized} employs a two-stage strategy to adapt diffusion into a general video-prediction model conditioned on text and actions, enabling large-scale driving simulation and planning. Pandora~\cite{xiang2024pandora} uses an instruction-tuned LLM to autoregressively steer a separate video-diffusion generator for explicit, goal-directed control, while Yume~\cite{mao2025yume} quantized camera motions into text tokens to guide a masked Video DiT, enabling autoregressive synthesis of dynamic 3D exploratory worlds.

To maintain geometric fidelity and long-horizon stability, recent methods couple explicit 3D priors with temporal-consistency modules in diffusion-based world models. At the geometric level, Geometry Forcing~\cite{wu2025geometry} aligns latent features with a geometric foundation model to inject explicit 3D priors, improving geometric consistency, while DeepVerse~\cite{chen2025deepverse} integrates visual and geometric prediction targets and introduces a geometry-aware memory to sustain consistent long-horizon generation. For temporal stability, VRAG~\cite{chen2025learning} proposed a Video Retrieval-Augmented Generation (RAG) framework that retrieves historical frames conditioned on a global state to stabilize autoregressive rollouts, StateSpaceDiffuser~\cite{savov2025statespacediffuser} combines Mamba with diffusion to alleviate long-term memory loss and content drift under short context windows, and InfinityDrive~\cite{guo2024infinitydrive} injects memory and adopts an adaptive loss within DiT, producing minute-scale driving videos with high fidelity, temporal consistency, and diverse content. Complementing these designs, LongDWM~\cite{wang2025longdwm} mitigates error accumulation in long-horizon video generation through distillation—where a fine-grained DiT learns continuous motion to guide a coarse model, whereas MiLA~\cite{wang2025mila} adopts a coarse-to-fine strategy that predicts sparse anchor frames and refines them during interpolation to improve temporal consistency and long-term fidelity. Finally, for dynamics and conditioning, Orbis~\cite{mousakhan2025orbis} employs a continuous-space flow-matching formulation that demonstrates greater robustness than discrete-token schemes for long-horizon rollouts, and DriVerse~\cite{li2025driverse} leverages multimodal trajectory prompts with latent motion alignment to synthesize long-horizon driving videos from a single image and navigation trajectories.

Sequential world models increasingly act as learned simulators, providing action-conditioned rollouts for policy evaluation and training. WorldGym~\cite{quevedo2025worldgym} and WorldEval~\cite{li2025worldeval} generate action-conditioned rollouts and use VLM-based critics for evaluation, while WorldEval further leverages latent action representations to drive a DiT-based synthesizer. RLVR-World~\cite{wu2025rlvr} fine-tunes world models with RL with Verifiable Rewards (RLVR), using explicit metrics to close the pretraining-task objective gap. For safety risk prediction, Guan~\etal\cite{guan2025world} presented a framework for autonomous driving accident prediction that augments data with a domain-informed world model and enhances spatiotemporal reasoning using graph and temporal convolutions.

Beyond diffusion, sequence models broadened the capacity for long-range consistency. Po~\etal\cite{po2025long} integrated block-wise state space models for long-term memory with local attention for short-term coherence, enabling video generation with sustained memory and consistent dynamics. S2-SSM~\cite{petri2025learning} employs a Mamba layer to model the independent evolution of object slots and a sparsity-regularized cross-attention mechanism to capture their causal interactions, enabling causal reasoning over the environment. 

\para{Spatial Latent Grid.}  Pretraining geometry-aligned spatial maps with self-supervised spatiotemporal objectives, the spatial latent grid paradigm preserves locality and enables efficient rollouts, multimodal fusion, and transferable planner-ready maps.

Building on this paradigm, structured-grid and physics-informed methods encode geometry and dynamics for controllable rollouts. PhyDNet~\cite{guen2020disentangling} disentangles physical priors expressed as partial differential equations from visual factors, improving prediction. ViDAR~\cite{Yang2024visual} unifies semantics, geometry, and dynamics through a pre-training task of point cloud forecasting and a latent rendering operator, enabling a scalable foundation for downstream autonomous driving tasks. FOLIAGE~\cite{liu2025foliage} models dynamics with an Accretive Graph Network and a Transformer-based predictor, executing rollouts on simulated data. Complementing these grid and physics lines, MindJourney~\cite{yang2025mindjourney} couples VLMs with a controllable world model to render egocentric rollouts along planned camera trajectories, enabling multi-view reasoning.

Building on grid-based representations, diffusion-based forecasting has become dominant for stable long-horizon generation. Within grid-centric predictors, DOME~\cite{gu2024dome} encodes observations into a continuous latent space and applies a spatiotemporal DiT for scene forecasting, Copilot4D~\cite{zhang2024copilot4d} tokenizes point clouds and couples a spatiotemporal Transformer with discrete diffusion to improve fidelity and coherence, and LidarDM~\cite{zyrianov2025lidardm} generates layout-conditioned static scenes, composes them with dynamic objects, integrating LiDAR simulation to produce controllable videos. For long-video generation, Vista~\cite{gao2024vista} adopts a two-stage large-scale training regime to produce controllable, high-fidelity, driving videos, and Delphi~\cite{ma2024unleashing} enforces long-horizon multiview consistency through shared noise and feature alignment and a failure-driven framework to synthesize targeted data for improved planning. To strengthen long-horizon stability, GEM~\cite{hassan2025gem} achieves controllable ego-vision generation through large-scale training and fine-grained control over motion, dynamics, and posture, Zhou~\etal\cite{zhou2025learning} maintains a persistent RGB-D 3D memory map to guide subsequent frames, and STAGE~\cite{wang2025stage} introduced hierarchical temporal feature transfer with multi-stage training.

\para{Decomposed Rendering Representation.} Scenes are decomposed into explicit primitives to synthesize view-consistent, simulatable trajectories over long horizons. Within this paradigm, GaussianWorld~\cite{zuo2025gaussianworld} models scene evolution as ego-motion, object dynamics, and newly observed regions, iteratively updating 3D Gaussian primitives to enable accurate and efficient dynamic perception. InfiniCube~\cite{lu2025infinicube} introduced a hybrid pipeline that combines voxel-based generation, video synthesis, and dynamic Gaussian reconstruction, enabling large-scale dynamic 3D driving scenes conditioned on HDmaps, bounding boxes, and text. Complementarily, Wu~\etal\cite{wu2025video} augmented a video world model with long-term spatial memory grounded in reconstructed geometry and an episodic memory, which together condition sequential generation for long-range consistency.

\subsubsection{Global Difference Prediction}\label{subsubsec:General_Global}
\textbf{Token Feature Sequence.}\quad For general-purpose world models, tokenized feature sequences support global prediction via masked and generative modeling, enabling parallel long-horizon rollouts with global constraints and multimodal conditioning. 

Within Joint-Embedding Predictive Architecture (JEPA)~\cite{LeCun2022Path}, V-JEPA~\cite{bardes2024revisiting} extends this architecture to video by predicting latent features of occluded spatiotemporal regions, learning generalizable representations for both appearance and motion without pixel reconstruction or contrastive learning. Building on this, V-JEPA 2~\cite{assran2025v} scales pretraining to large-scale Internet videos with larger models and incorporates  limited robot interaction data for post-training, transferring to robotic planning. V-JEPA 2.1~\cite{mur2026v} further modifies this recipe with dense prediction over both visible and masked tokens, deep self-supervision, and image-video tokenizers, improving local spatiotemporal features while preserving global prediction and embodied planning capability. DINO-Foresight~\cite{karypidis2026dino} similarly treats pretrained DINO/VFM features as semantic latent tokens, using a masked feature Transformer to forecast future driving-scene representations that can be decoded by segmentation, depth, and surface-normal heads. AD-L-JEPA~\cite{zhu2025ad} adapts JEPA to BEV LiDAR, predicting masked embeddings in a self-supervised manner. Beyond JEPA-style prediction, WorldDreamer~\cite{wang2024worlddreamer} frames world modeling as masked visual sequence prediction to learn physics and motion for diverse video generation and editing, while MaskGWM~\cite{ni2025maskgwm} combines diffusion with masked feature reconstruction and a dual-branch masking strategy to improve long-horizon consistency and generalization.

In parallel, diffusion-based methods have become central to global-difference modeling. Sora~\cite{brooks2024video} represents video as unified spacetime patches and uses a DiT to generate long, coherent sequences at scale. Extending this line toward Physical AI, Cosmos 3~\cite{agarwal2026cosmos} unifies language, image, video, audio, and action sequences within a mixture-of-transformers architecture, positioning omnimodal world models as general-purpose backbones for understanding, generation, simulation, and action prediction. ForeDiff~\cite{zhang2025consistent} decouples conditioning from denoising by adding a deterministic predictive stream and employing a pretrained predictor to guide generation, improving accuracy and consistency. For domain-specific synthesis, AirScape~\cite{zhao2025airscape} introduces an aerial video-intention dataset, applies supervised fine-tuning for controllability, and leverages VLMs to impose spatiotemporal constraints; MarsGen~\cite{li2025martian} builds a multimodal Mars dataset from NASA's sparse rover stereo imagery using geometric foundation models, then trains a controllable generator to produce visually realistic, geometry-consistent Martian videos. In clinical guidance, EchoWorld~\cite{yue2025echoworld} proposes a motion-aware world model for echocardiography probe control, pretraining on region- and motion-outcome prediction and fine-tuning attention to fuse visual and motion cues for precise guidance.

\para{Spatial Latent Grid.}  Spatial-grid models forecast voxel grids in parallel and fuse multi-view visual features into a unified map, learning a general-purpose world model.

Recent work converges on unified scene understanding and future prediction. UniFuture~\cite{liang2025seeing} couples Dual Latent Sharing with multi-scale latent interaction to jointly model appearance and depth in future driving scenes, and HERMES~\cite{zhou2025hermes} integrates multiview BEV features into an LLM with world queries that link scene understanding to future prediction within a single framework. BEVWorld~\cite{zhang2024bevworld} maps images and LiDAR into a compact BEV latent space through a unified tokenizer and applies a latent BEV diffusion model for synchronized multimodal forecasting. Progress in grid and occupancy forecasting includes differentiable raycasting with a proxy reformulation for sensor agnostic motion learning by Khurana~\etal\cite{khurana2022differentiable,khurana2023point} and LiDAR to range images 3D spatiotemporal convolutions by Mersch~\etal\cite{mersch2022self}. Cam4DOcc~\cite{ma2024cam4docc} established the first vision-only benchmark with an E2E 3D CNN baseline, and Liu~\etal\cite{liu2025towards} enhanced cross-task transfer through high-ratio compression and latent flow matching.

On the generative front, tokenized 4D representations enable controllable scene synthesis. OccSora~\cite{wang2024occsora} uses a 4D tokenizer to derive compact representations for trajectory conditioned diffusion, and DynamicCity~\cite{bian2025dynamiccity} encodes 4D occupancy as HexPlanes representations with a VAE and employs a conditional DiT for high fidelity controllable dynamics. Fidelity and consistency improve through decoupling ego-motion with scene evolution in COME~\cite{shi2025come}, physics-informed constraints in DrivePhysica~\cite{yang2024physical}, cross-view point map alignment in Liu~\etal\cite{liu2025geometry}, and photometric warping-based supervision in PosePilot~\cite{jin2025posepilot}. For controllable conditioning, DriveDreamer 2~\cite{zhao2025drivedreamer} translates prompts into agent trajectories and HDMaps for customizable video generation, EOT-WM~\cite{zhu2025other} encodes ego and surrounding trajectories as trajectory videos for trajectory consistent synthesis, and ORV~\cite{yang2025orv} uses 4D semantic occupancy sequences to guide action conditioned video with S2R transfer. AETHER~\cite{team2025aether} unifies dynamic 4D reconstruction, action-conditioned video prediction, and vision-based planning under training on synthetic 4D data and achieves zero-shot generalization to real-world scenarios.

\para{Decomposed Rendering Representation.} This paradigm performs global prediction by combining explicit 3D structure with video generative priors.

A trend is combining video generation with Gaussian Splatting. DriveDreamer4D~\cite{zhao2025drivedreamer4d} exploits complex driving trajectories such as lane changes to guide video synthesis and optimize a 4DGS model, which enhances reconstruction fidelity and spatiotemporal coherence from novel viewpoints. ReconDreamer~\cite{ni2025recondreamer} introduces an online restoration module together with progressive data reuse to correct artifacts in Gaussian-rendered views and enables reliable reconstruction of large-scale trajectories. MagicDrive3D~\cite{gao2024magicdrive3d} generates multi-view street scenes conditioned on BEV maps, 3D boxes, and text, and further converts the outputs into full 3D environments through fault-tolerant GS. In contrast, implicit-field methods replace GS with continuous neural representations. UnO~\cite{agro2024uno} leverages future point clouds to learn a NeRF-style 4D occupancy field, which allows annotation-free prediction and achieves strong transfer performance beyond supervised baselines in point-cloud forecasting.

In summary, decision coupling specifies how strongly a world model is tied to a particular control objective, temporal modeling governs how predictions unfold over time, and spatial representation determines the level of geometric detail and locality. Different combinations of these axes induce characteristic trade-offs: decision-coupled RSSM-style models with global latent vectors favor sample efficiency and fast rollouts at the cost of visual detail; token-based simulators, especially with global prediction, improve fidelity and semantic richness but incur higher computational and data demands; grid-based and decomposed-rendering models provide strong geometric and multi-view coherence, yet remain challenging to deploy under strict latency and memory budgets.

\section{Data Resources \& Metrics}\label{sec:data}
World models in embodied AI are required to address diverse tasks spanning manipulation, navigation, and autonomous driving, requiring heterogeneous resources and rigorous evaluation. Accordingly, we present Data Resources in \secref{subsec:Data_Resources} and Metrics in \secref{subsec:metrics}, focusing on the most widely adopted platforms and evaluation measures as a unified foundation for cross-domain assessment.

\subsection{Data Resources}\label{subsec:Data_Resources}
To meet the diverse demands of embodied AI, we categorize data resources into four categories: Simulation Platforms, Interactive Benchmarks, Offline Datasets, and Real-world Robot Platforms, as detailed in the following subsections. \tabref{tab:dataset} provides a comprehensive overview of these resources.

\begin{table*}[!t]
\centering
\setlength{\tabcolsep}{3.mm}
\caption{\textbf{An overview of data resources for training and evaluating embodied world models.}}
\label{tab:dataset}
\renewcommand{\arraystretch}{1.2}
\resizebox{\linewidth}{!}{%
\begin{tabular}{c l  c c c c c c }
\toprule[1.5pt]
\textbf{Category} & \textbf{Name} & \textbf{Year} & \textbf{Task}  &  \textbf{Input} & \textbf{Domain} & \textbf{Scale} & \textbf{Protocol\textsuperscript{\protect\hyperlink{note:Protocol}{1}}}   \\
\midrule
\multirow{5}{*}{\rotatebox{90}{\textbf{Platform}}} 
& MuJoCo\cite{todorov2012mujoco} & 2012 & Continuous control    & Proprio. & Sim & - & - \\
& CARLA\cite{dosovitskiy2017carla} & 2017 & Driving simulation  & RGB-D/Seg/LiDAR/Radar/GPS/IMU & Sim & - & \cmark \\
& Habitat\cite{savva2019habitat} & 2019 & Embodied navigation  & RGB-D/Seg/GPS/Compass & Sim & - & \cmark  \\
& Isaac Gym\cite{makoviychuk2021isaac} & 2021 & continuous control  & Proprio.  & Sim & - & -  \\
& Isaac Lab\cite{mittal2023orbit} & 2023 & Robot learning suites  & RGB-D/Seg/LiDAR/Proprio.   & Sim & - & -  \\
\midrule
\multirow{6}{*}{\rotatebox{90}{\textbf{Benchmark}}} 
 & Atari\cite{bellemare2013arcade} & 2013 & Discrete-action game & RGB/State & Sim & 55\!+\ Games & \cmark  \\
 & DMC\cite{tassa2018deepmind} & 2018   & Continuous control & RGB/Proprio. & Sim & 30\!+\ Tasks & \cmark  \\
 & Meta-World\cite{yu2020meta} & 2019 & Multi-task manipulation  & RGB/Proprio. & Sim & 50 tasks &  \cmark  \\
 & RLBench\cite{james2020rlbench} & 2020 & Robotic manipulation  & RGB-D/Seg/Proprio. & Sim & 100 tasks & \cmark  \\
 & nuPlan\cite{caesar2021nuplan} & 2021 & Driving planning & RGB/LiDAR/Map/Proprio. & Real & 1.5k hours & \cmark \\
 & LIBERO\cite{liu2023libero} & 2023 & Lifelong manipulation  & RGB/Text/Proprio. & Sim & 130 tasks & \cmark  \\
\midrule
\multirow{10}{*}{\rotatebox{90}{\textbf{Dataset}}} 
 & SSv2\cite{goyal2017something} & 2018 & Video-action understanding  & RGB/Text & Real & 220k videos & 169k/24k/27k  \\ 
 & nuScenes\cite{caesar2020nuscenes} & 2020 & Driving perception  & RGB/LiDAR/Radar/GPS/IMU & Real &  1k scenes & 700/150/150  \\
 & Waymo\cite{sun2020scalability} & 2020 & Driving perception & RGB/LiDAR & Real & 1.15k scenes & 798/202/150  \\
 & HM3D\cite{ramakrishnan2021habitatmatterport} & 2021 & Indoor navigation & RGB-D & Real & 1k scenes & 800/100/100  \\
 & RT-1\cite{brohan2022rt} & 2022 & Real-robot manipulation  & RGB/Text & Real & 130k\!+\ trajectories & - \\
 & Occ3D\cite{tian2023occ3d} & 2023 & 3D occupancy  & RGB/LiDAR & Real & 1.9k scenes & 600/150/150; 798/202/-  \\ 
 & OXE\cite{o2024open} & 2024 & Cross-embodiment pretraining & RGB-D/LiDAR/Text & Real & 1M\!+\ trajectories & - \\
 & OpenDV\cite{yang2024generalized} & 2024 & Driving video pretraining  & RGB/Text & Real & 2k\!+\ hours & - \\
 & VideoMix22M\cite{assran2025v} & 2025 & Video pretraining  & RGB & Real & 22M\!+\ samples & - \\
 & VisionMix163M\cite{mur2026v} & 2026 & Image-video pretraining  & RGB & Real & 163M\!+\ samples & - \\
\midrule
\multirow{3}{*}{\rotatebox{90}{\textbf{Robot}}}
 & Franka Emika\cite{haddadin2022franka} & 2022 & Manipulation & Proprio. & Real & - & -  \\
 & Unitree Go1~\cite{unitree_go1_web} & 2021 & Quadruped locomotion & RGB-D/LiDAR/Proprio. & Real & - & -  \\
 & Unitree G1\cite{unitree_g1_web} & 2024 & Humanoid manipulation & RGB-D/LiDAR/Proprio./Audio & Real & - & -  \\ 
\bottomrule[1.5pt]
\end{tabular}}
\begin{tablenotes}[flushleft]\fontsize{6pt}{7pt}\selectfont
    \item \hypertarget{note:Protocol}{}\textsuperscript{1} \textbf{Protocol}: For interactive  benchmarks, a check mark (\cmark) indicates available evaluation protocols. For datasets, it indicates official data splits are provided.
\end{tablenotes}
\end{table*}

\subsubsection{Simulation Platforms}\label{subsubsec:Simulation_Platforms}
Simulation platforms provide controllable and scalable virtual environments for training and evaluating world models.
\begin{itemize}[leftmargin=*]
    \item \textbf{MuJoCo}~\cite{todorov2012mujoco} is a customizable physics engine widely adopted for its efficient robotic simulation of articulated systems and contact dynamics in robotics and control research.

    \item \textbf{NVIDIA Isaac} is an E2E, GPU-accelerated simulation stack that comprises Isaac Sim, Isaac Gym~\cite{makoviychuk2021isaac}, and Isaac Lab~\cite{mittal2023orbit}. It offers photorealistic rendering and large-scale RL capabilities.

    \item \textbf{CARLA}~\cite{dosovitskiy2017carla} is an open-source simulator based on Unreal Engine for urban autonomous driving, providing realistic rendering, diverse sensors, and closed-loop evaluation protocols.
    
    \item \textbf{Habitat}~\cite{savva2019habitat} is a high-performance simulator for embodied AI, specializing in photorealistic 3D indoor navigation.

\end{itemize}
\subsubsection{Interactive Benchmarks}\label{subsubsec:Interactive_Benchmarks}
Interactive benchmarks offer standardized task suites and protocols for reproducible closed-loop evaluation of world models.

\begin{itemize}[leftmargin=*]
    \item \textbf{DeepMind Control (DMC)}~\cite{tassa2018deepmind} is a standard MuJoCo-based suite for control tasks, offering a consistent basis for comparing agents that learn from state or pixel-based observations.
    
    \item \textbf{Atari}~\cite{bellemare2013arcade} is a suite of pixel-based, discrete-action games for evaluating agent performance. The Atari100k~\cite{kaiser2020model} specifically assesses sample efficiency by limiting interaction to 100k steps.
    
    \item \textbf{Meta-World}~\cite{yu2020meta} is a benchmark for multi-task and meta-RL, featuring \num{50} diverse robotic manipulation tasks with a Sawyer arm in MuJoCo under standardized evaluation protocols.
    
    \item \textbf{RLBench}~\cite{james2020rlbench} offers \num{100} simulated tabletop manipulation tasks with sparse rewards and rich, multi-modal observations, designed to test complex skills and rapid adaptation.

    \item \textbf{LIBERO}~\cite{liu2023libero} is a benchmark for lifelong robotic manipulation, providing \num{130} procedurally generated tasks and human demonstrations to evaluate sample-efficient and continual learning.

    \item \textbf{nuPlan}~\cite{caesar2021nuplan} is a planning benchmark for autonomous driving, using a lightweight closed-loop simulator and over \SI{1500}{\hour} of real-world driving logs to evaluate long-horizon performance.
\end{itemize}

\subsubsection{Offline Datasets}
\label{subsubsec:Offline_Datasets}
Offline datasets are large-scale, pre-collected trajectories that eliminate interactive rollouts and provide a foundation for reproducible evaluation and data-efficient pretraining of world models.
\begin{itemize}[leftmargin=*]
    \item \textbf{RT-1}~\cite{brohan2022rt} is a real-world dataset for robot learning, collected over \num{17} months with \num{13} Everyday Robots mobile manipulators. It contains \num{130000} demonstrations spanning more than \num{700} tasks, pairing language instructions and image observations with discretized 11-DoF actions for the arm and mobile base.

    \item \textbf{Open X-Embodiment (OXE)}~\cite{o2024open} is a corpus aggregating \num{60} sources from \num{21} institutions, spanning \num{22} robot embodiments, \num{527} skills, and over one million trajectories in a unified format for cross-embodiment training. Models trained on OXE demonstrate strong transfer beyond single-robot baselines, underscoring the effectiveness of cross-platform data sharing.

    \item \textbf{Habitat-Matterport 3D (HM3D)}~\cite{ramakrishnan2021habitatmatterport} is a large-scale dataset of \num{1000} indoor reconstructions with \SI{112500}{\meter\squared} navigable area, substantially expanding the scope and diversity of embodied AI simulation. Released for the Habitat platform, it provides the necessary metadata and resources for seamless use.
    
    \item \textbf{nuScenes}~\cite{caesar2020nuscenes} is a large-scale multimodal driving dataset with a 360-degree sensor suite comprising six cameras, five radars, one LiDAR, and GPS/IMU. It contains \num{1000} twenty-second scenes collected in Boston and Singapore with dense 3D annotations for \num{23} classes and HDMaps, providing a core benchmark for multimodal fusion and long-horizon prediction.

    \item \textbf{Waymo}~\cite{sun2020scalability} is a multimodal autonomous driving benchmark with \num{1150} twenty-second scenes at \SI{10}{\hertz} from San Francisco, Phoenix, and Mountain View. It includes five LiDARs and five cameras, with about \num{12} million 3D and 2D annotations, making it a large-scale resource for modeling traffic dynamics.

    \item \textbf{Occ3D}~\cite{tian2023occ3d} defines 3D occupancy prediction from surround-view images, providing voxel labels that distinguish free, occupied, and unobserved states. Occ3D-nuScenes contains about \num{40000} frames at \SI{0.4}{\meter} resolution, while Occ3D-Waymo offers about \num{200000} frames at \SI{0.05}{\meter}. This voxel-level supervision enables holistic scene understanding beyond bounding boxes.

    \item \textbf{Something-Something~v2 (SSv2)}~\cite{goyal2017something} is a video dataset for fine-grained action understanding. It contains \num{220847} clips across \num{174} categories, collected from crowd workers following textual prompts (\eg, Putting something into something) with splits of \num{168913} train, \num{24777} val, and \num{27157} test videos.
    
    \item \textbf{OpenDV}~\cite{yang2024generalized} is the largest large-scale video-text dataset for autonomous driving, proposed by GenAD, which supports video prediction and world-model pretraining. It contains \num{2059} hours and \num{65.1} million frames from YouTube and seven public datasets, covering over \num{40} countries and \num{244} cities. The dataset provides command and context annotations to enable language- and action-conditioned prediction and planning.

    \item \textbf{VideoMix22M}~\cite{assran2025v} is a large-scale dataset introduced with V-JEPA 2 for self-supervised pretraining. It scales from \num{2} million to \num{22} million samples, drawn from YT-Temporal-1B~\cite{zellers2022merlot}, HowTo100M~\cite{miech2019howto100m}, Kinetics~\cite{carreira2019short}, SSv2, and ImageNet~\cite{deng2009imagenet}. The largest source, YT-Temporal-1B, is curated with retrieval-based filtering to suppress noise, while ImageNet images are converted into static video clips for consistency.

    \item \textbf{VisionMix163M}~\cite{mur2026v} extends VideoMix-style self-supervised pretraining by combining SSv2, Kinetics, HowTo100M, YT-Temporal-1B, and LVD-142M image data. It replaces the ImageNet subset used in V-JEPA 2 with a much larger curated image source and rebalances video sampling toward motion-rich content to support dense and temporally consistent visual features.

\end{itemize}

\subsubsection{Real-world Robot Platforms}\label{subsubsec:Real-world_Robot_Platforms}
Real-world robot platforms provide physical embodiments for interaction, enabling closed-loop evaluation, high-fidelity data collection, and S2R validation under real-world constraints.
\begin{itemize}[leftmargin=*]
    \item \textbf{Franka Emika}~\cite{haddadin2022franka} is a 7-DoF collaborative robot arm with joint torque sensors for precise force control. Through the control interface, it supports \SI{1}{\kilo\hertz} torque control for contact-rich tasks, while its ROS integration makes it a versatile platform.  
    
    \item \textbf{Unitree Go1}~\cite{unitree_go1_web} is a cost-effective and widely adopted quadrupedal robot equipped with a panoramic depth-sensing suite, onboard computing of 1.5~TFLOPS, and a maximum speed of \SI{4.7}{\meter\per\second}, establishing it as a standard platform for locomotion and embodied-AI research.

    \item \textbf{Unitree G1}~\cite{unitree_g1_web} is a compact humanoid robot for research, offering up to \num{43}-DoF and knee torques of \num{120}~N$\cdot$m, with integrated 3D LiDAR and depth cameras. With multimodal sensing, onboard compute, ROS support, and swappable batteries, this low-cost platform provides a practical real-robot testbed for training and evaluating embodied world models.    
\end{itemize}

\subsection{Metrics}\label{subsec:metrics}
Metrics evaluate the capability of world models to capture dynamics, generalize to unseen scenarios, and scale with additional resources. We organize them into three abstraction levels: \secref{subsubsec:Pixel} \textbf{Pixel Prediction Quality},  \secref{subsubsec:State} \textbf{State-level Understanding}, and \secref{subsubsec:Task} \textbf{Task Performance}, representing a progression from low-level signal fidelity to high-level goal attainment.
\subsubsection{Pixel Generation Quality}\label{subsubsec:Pixel}
At the most fundamental level, world models are evaluated by their ability to reconstruct sensory inputs and generate realistic sequences. Metrics assess image fidelity, temporal consistency, and perceptual similarity, providing quantitative measures of the extent to which models capture raw environmental dynamics.

\para{Fr\'{e}chet Inception Distance (FID)~\cite{heusel2017gans}.} FID is a metric for assessing the realism and diversity of generated images. It compares real and generated image distributions in the feature space of an ImageNet-pretrained Inception-v3~\cite{szegedy2016rethinking}, modeling embeddings as Gaussians with means $\boldsymbol{\mu}_x,\boldsymbol{\mu}_y$ and covariances $\boldsymbol{\Sigma}_x,\boldsymbol{\Sigma}_y$. Defined as
\begin{equation}
    \label{eq:FID}
    \operatorname{FID}(x,y) = \lVert \boldsymbol{\mu}_x - \boldsymbol{\mu}_y \rVert_2^2 + \operatorname{Tr}\left( \boldsymbol{\Sigma}_x + \boldsymbol{\Sigma}_y - 2(\boldsymbol{\Sigma}_x \boldsymbol{\Sigma}_y)^{1/2} \right),
\end{equation}
a lower FID denotes a closer alignment between real and generated distributions. By comparing the first and second moments, it penalizes fidelity loss (mean shift) and mode collapse (covariance mismatch), offering a holistic measure of generative performance.

\para{Fr\'{e}chet Video Distance (FVD)~\cite{unterthiner2018towards}.} FVD extends FID to videos, evaluating both per-frame quality and temporal consistency. It replaces the image-based Inception network with an I3D~\cite{carreira2017quo} pretrained on Kinetics-400~\cite{kay2017kinetics}. Using the same Fr\'echet distance as \equref{eq:FID} on motion-aware features, FVD yields a holistic video quality score. A lower value indicates a closer alignment of distributions in appearance and dynamics while penalizing temporal artifacts like unnatural motion or flickering.

\para{Structural Similarity Index Measure (SSIM)~\cite{wang2004image}.} SSIM is a perceptual metric for image quality that compares luminance, contrast, and structure between a generated image and its reference. For two  patches $x$ and $y$ with means $\boldsymbol{\mu}_x$, $\boldsymbol{\mu}_y$, variances $\boldsymbol{\Sigma}_x^2,\boldsymbol{\Sigma}_y^2$, and covariance $\boldsymbol{\Sigma}_{xy}$, SSIM is defined as
\begin{equation}
  \operatorname{SSIM}(x,y)=\frac{(2\boldsymbol{\mu}_x\boldsymbol{\mu}_y+C_1)(2\boldsymbol{\Sigma}_{xy}+C_2)}{(\boldsymbol{\mu}_x^2+\boldsymbol{\mu}_y^2+C_1)(\boldsymbol{\Sigma}_{x}^2+\boldsymbol{\Sigma}_{y}^2+C_2)}.
\end{equation}
The final score is obtained by averaging SSIM over sliding windows, and values closer to $1$ indicate higher similarity.

\para{Peak Signal-to-Noise Ratio (PSNR)~\cite{hore2010image}.} PSNR measures pixel-wise distortion between a reconstruction and its reference. Let the mean-squared error (MSE) over $N$ pixels be
\begin{equation}
  \operatorname{MSE}=\frac{1}{N}\sum_{i=1}^{N}\left(x_i-y_i\right)^2,
\end{equation}
and let $\operatorname{MAX}$ denote the maximum possible pixel value(\eg, $255$ for RGB or $1$ for normalized images). Then
\begin{equation}
  \operatorname{PSNR}(x,y)= 10\cdot\log_{10}\left(\frac{\mathrm{MAX}^2}{\mathrm{MSE}}\right).
\end{equation}
Higher PSNR values indicate lower distortion and greater fidelity. 

\para{Learned Perceptual Image Patch Similarity  (LPIPS)~\cite{zhang2018unreasonable}.} LPIPS is a metric that correlates with human judgments by comparing features extracted from pretrained networks. Let $\hat{f}^l_x$ and $\hat{f}^l_y$ denote the unit-normalized activations at layer $l$ for inputs $x$ and $y$, and $w_l$ the channel-wise weights. LPIPS is defined as
\begin{equation}
  \operatorname{LPIPS}(x,y) = \sum_{l} \frac{1}{H_l W_l} \sum_{h,w} \left\| w_l \odot \big(\hat{f}_{h,w,x}^{l} - \hat{f}_{h,w,y}^l\big)\right\|_2^2.
\end{equation}
Lower LPIPS values imply greater similarity, offering enhanced fidelity compared to pixel-based metrics and robustness against distortions.

\para{VBench~\cite{huang2024vbench}.} VBench is a comprehensive benchmark for video generation that assesses performance across 16 dimensions grouped into two categories: Video Quality (\eg, subject consistency, motion smoothness) and Video-Condition Consistency (\eg, object class, human action). It provides carefully curated prompt suites and large-scale human preference annotations to ensure strong perceptual alignment, thereby enabling fine-grained evaluation of model capabilities and limitations.

\subsubsection{State-level Understanding}\label{subsubsec:State}
Beyond pixel fidelity, state-level understanding assesses whether models capture objects, layouts, and semantics, and can predict their evolution. Metrics span semantic, BEV, and 3D segmentation, detection, occupancy, geometry, and trajectory accuracy, emphasizing structural understanding beyond appearance.

\para{mean Intersection over Union (mIoU).} mIoU evaluates semantic segmentation by averaging the Intersection over Union (IoU) across classes. For class $c$,
\begin{equation}
  \operatorname{IoU}=\frac{\operatorname{TP}}{\operatorname{TP}+\operatorname{FP}+\operatorname{FN}},
\end{equation}
where TP, FP, and FN denote the true positives, false positives, and false negatives. IoU quantifies overlap with the ground truth while penalizing segmentation errors. The dataset-level score is
\begin{equation}
  \operatorname{mIoU}=\frac{1}{\left| C \right|}\sum_{c\in C}\operatorname{IoU}_c.
\end{equation}
A higher mIoU reflects more precise semantic scene understanding.

\para{mean Average Precision (mAP).} mAP evaluates detection and instance segmentation by averaging per-class Average Precision (AP). For a class $c$ at IoU threshold $\tau$, predictions are ranked by confidence and matched one-to-one with ground truths when $\mathrm{IoU}\ge\tau$ with unmatched predictions counted as FP and unmatched ground truths as FN. Precision and recall are
\begin{equation}
  \operatorname{Precision}=\frac{\operatorname{TP}}{\operatorname{TP+FP}},\quad \operatorname{Recall}=\frac{\operatorname{TP}}{\operatorname{TP}+\operatorname{FN}}.
\end{equation}
Let $P_{c,\tau}(r)$ denote the precision-recall envelope obtained via monotonic interpolation. The AP for class $c$ at threshold $\tau$ is
\begin{equation}
  \operatorname{AP}_{c,\tau}=\int_{0}^{1} P_{c,\tau}(r) \mathrm{d}r.
\end{equation}
mAP averages AP across classes and thresholds $T$:
\begin{equation}
  \operatorname{mAP}=\frac{1}{\left| C \right|}\sum_{c\in C}\left( \frac{1}{\left| T \right|} \sum_{\tau\in T}\operatorname{AP}_{c,\tau}\right).
\end{equation}
A higher mAP indicates better instance recognition, more accurate localization, and more calibrated confidence estimates.

\para{Displacement Error.} Displacement error metrics assess state-level understanding by measuring spatial accuracy for keypoints, object centers, and trajectory waypoints. The L2 trajectory error computes the Euclidean distance between predicted and ground-truth waypoints. Common variants include Average Displacement Error (ADE), which calculates the average displacement, and Final Displacement Error (FDE), which measures the displacement at the final step. Lower values indicate more accurate localization.

\para{Chamfer Distance (CD)~\cite{fan2017point}.} CD quantifies geometric similarity between a prediction $S_1$ and ground truth $S_2$ by summing squared nearest-neighbor distances across the two sets:
\begin{equation}
   \operatorname{CD}(S_{1},S_{2})=\sum_{x\in S_{1}}\min_{y\in S_{2}}\left\|x-y\right\|_{2}^{2}+\sum_{y\in S_{2}}\min_{x\in S_{1}}\left\|x-y\right\|_{2}^{2}.
\end{equation}
Unlike pixel-level metrics, CD captures surfaces, occupancy, BEV, and 3D structures, and its differentiability enables use as both a training loss and an evaluation metric that complements IoU.

\subsubsection{Task Performance}\label{subsubsec:Task}
Ultimately, the value of a world model lies in supporting effective decision-making, with task-level metrics assessing goal achievement under safety and efficiency constraints in embodied settings.

\para{Success Rate (SR).} SR quantifies performance as the fraction of evaluation episodes that satisfy a predefined success condition. In navigation and manipulation, the condition is typically binary, such as reaching a target or placing an object correctly. In autonomous driving, the requirement is stricter, demanding route completion without collisions or major violations. The final SR is reported as the average of binary outcomes across all test episodes.

\para{Sample Efficiency (SE).} SE quantifies the samples needed to reach target performance. It is evaluated by fixed-budget benchmarks (\eg, Atari-100k), data-performance curves, or in robotics by the demonstrations required to achieve a given success rate.

\para{Reward.} In RL, the reward is a signal $r_t$ at timestep $t$. The goal is to maximize the discounted return $G_t=\sum_{k=0}^\infty \gamma^k r_{t+k+1}$. Results are reported as cumulative reward or average return, often with normalization for cross-task comparison. 

\para{Collision.} Safety is evaluated with collision-based metrics. The primary measure, collision rate, is the proportion of evaluation episodes with at least one collision and is common in indoor navigation. In autonomous driving, exposure-normalized variants are used, such as collisions per kilometer or collisions per hour.

\section{Performance Comparison}\label{sec:performance}
Given the proliferation of world-model variants and heterogeneous metrics, we organize comparisons by task objectives and rely on standardized benchmarks, reporting concise tables that highlight each method's strengths and limitations.

\subsection{Pixel Generation}\label{subsec:pixel-generation}
\textbf{Generation on nuScenes.}\quad Driving video generation is treated as a world-modeling task that synthesizes plausible scene dynamics in fixed-length clips. Typical protocols produce short sequences and evaluate quality with \emph{FID} for appearance fidelity and \emph{FVD} for temporal consistency. For a fair comparison on the nuScenes validation split, recent approaches have achieved remarkable progress, as shown in \tabref{tab:Generation}. DrivePhysica delivers the best visual fidelity, while MiLA achieves the strongest temporal coherence, together establishing new state-of-the-art performance.
\begin{table}[t]
  \centering
  \caption{\textbf{Performance comparison of video generation on the nuScenes.}}
  \label{tab:Generation}
  \renewcommand{\arraystretch}{1.1}
  \resizebox{\columnwidth}{!}{%
  \begin{tabular}{l|c|ccc}
    \specialrule{1.5pt}{1pt}{0pt} 
    \rowcolor{gray!15}
    \textbf{Method} & \textbf{Pub.}& \textbf{Resolution} & \textbf{FID$\downarrow$} & \textbf{FVD$\downarrow$}  \\
    \specialrule{0.5pt}{0pt}{0.5pt}\specialrule{0.5pt}{0.5pt}{1pt}

      MagicDrive3D\cite{gao2024magicdrive3d}  & arXiv'24 & $224 \times 400 $
      & 20.7 & 164.7 \\

      Delphi\cite{ma2024unleashing}           & arXiv'24 & $512 \times 512 $
      & 15.1 & 113.5 \\

      Drive-WM\cite{wang2024driving}          & CVPR'24 & $192 \times 384$ 
      & 15.8 & 122.7 \\
      GenAD\cite{yang2024generalized}         & CVPR'24 & $256 \times 448$ 
      & 15.4 & 184.0 \\

      DriveDreamer\cite{wang2024drivedreamer} & ECCV'24 & $128 \times 192$ 
      & 52.6 & 452.0 \\

      Vista\cite{gao2024vista}                & NeurIPS'24 & $576 \times 1024$ 
      & 6.9 & 89.4 \\

      DrivePhysica\cite{yang2024physical}     & arXiv'24 & $256 \times 448 $
      & \textbf{4.0} & 38.1 \\

      DrivingWorld\cite{hu2024drivingworld}   & arXiv'24 & $512 \times 1024$ 
      & 7.4 & 90.9 \\

      DriveDreamer-2\cite{zhao2025drivedreamer}& AAAI'25 & $256 \times 448 $
      & 11.2 & 55.7 \\

      UniFuture\cite{liang2025seeing}         & arXiv'25 & $320 \times 576 $
      & 11.8 & 99.9 \\

      MiLA\cite{wang2025mila}                &  arXiv'25 & $360 \times 640 $
      & 4.1 & \textbf{14.9} \\

      GeoDrive\cite{chen2025geodrive}         & arXiv'25 & $480 \times 720 $
      & 4.1 & 61.6 \\

      LongDWM\cite{wang2025longdwm}          &  arXiv'25 & $480 \times 720 $
      & 12.3 & 102.9 \\

      MaskGWM\cite{ni2025maskgwm}            & CVPR'25  & $288 \times 512 $
      & 8.9 & 65.4 \\
      GEM\cite{hassan2025gem}                &  CVPR'25 & $576 \times 1024 $
      & 10.5 & 158.5 \\

      Epona\cite{zhang2025epona}              & ICCV'25 & $512 \times 1024 $ 
      & 7.5 & 82.8 \\

      STAGE\cite{wang2025stage}               & IROS'25 & $512 \times 768 $
      & 11.0 & 242.8 \\

      DriVerse\cite{li2025driverse}          &  ACMMM'25 & $480 \times 832 $
      & 18.2 & 95.2 \\

    \specialrule{1.5pt}{1pt}{0pt}
  \end{tabular}}
\end{table} 

\subsection{Scene Understanding}
\label{subsec:scenes_understanding}
\textbf{4D Occupancy Forecasting on Occ3D-nuScenes.}\quad 4D occupancy forecasting is treated as a representative world modeling task. Given \SI{2}{\second} of past 3D occupancy, models predict the subsequent \SI{3}{\second} of scene dynamics. Evaluation follows the Occ3D-nuScenes protocol and reports \emph{mIoU} and per horizon \emph{IoU}. As summarized in \tabref{tab:nuscenes_occ}, we compare methods by input modality, auxiliary supervision, and ego trajectory usage to reveal design choices for spatiotemporal forecasting. Methods using occupancy inputs outperform camera-only variants, and adding auxiliary supervision with a GT ego trajectory further mitigates performance decay at 2-3 s. Among all methods, COME (with GT ego) achieves the best average mIoU and per-horizon IoU.

\begin{table*}[t]
\centering
\caption{\textbf{Performance comparison of 4D occupancy forecasting on the Occ3D-nuScenes benchmark\textsuperscript{\protect\hyperlink{note:4DOcc}{1}}.}}
\label{tab:nuscenes_occ}
\renewcommand{\arraystretch}{1.1}
\resizebox{\linewidth}{!}{%
\begin{tabular}{l||ccc||C{.03\textwidth}C{.03\textwidth}C{.03\textwidth}C{.03\textwidth}C{.03\textwidth}C{.03\textwidth}C{.03\textwidth}C{.03\textwidth}C{.03\textwidth}C{.03\textwidth}}
\specialrule{1.5pt}{0pt}{0pt}
 & \multirow{2}{*}{\textbf{Input}} & \multirow{2}{*}{\textbf{Aux. Sup}} & \multirow{2}{*}{\textbf{Ego traj.}} 
 & \multicolumn{5}{c}{\textbf{mIoU (\%)} $\uparrow$} & \multicolumn{5}{c}{\textbf{IoU (\%) $\uparrow$}} \\
 
\multirow{-2}{*}{\textbf{Method}} & & & & \textbf{Recon.} & \textbf{1s} & \textbf{2s} & \textbf{3s} & \cellcolor{gray!20}\textbf{Avg.} & \textbf{Recon.} & \textbf{1s} & \textbf{2s} & \textbf{3s} & \cellcolor{gray!20}\textbf{Avg.} \\

\specialrule{0.5pt}{.5pt}{0.5pt}\specialrule{0.5pt}{0.5pt}{.5pt}

Copy \& Paste\textsuperscript{\protect\hyperlink{note:Copy_Paste}{2}} & Occ & None & Pred. &
66.38 & 14.91 & 10.54 & 8.52 & \cellcolor{gray!20}11.33 &
62.29 & 24.47 & 19.77 & 17.31 & \cellcolor{gray!20}20.52 \\

OccWorld-O\cite{zheng2024occworld} & Occ & None & Pred. &
66.38 & 25.78 & 15.14 & 10.51 & \cellcolor{gray!20}17.14 &
62.29 & 34.63 & 25.07 & 20.18 & \cellcolor{gray!20}26.63 \\

OccLLaMA-O\cite{wei2024occllama}  & Occ & None & Pred. &
75.20 & 25.05 & 19.49 & 15.26 & \cellcolor{gray!20}19.93 &
63.76 & 34.56 & 28.53 & 24.41 & \cellcolor{gray!20}29.17 \\

RenderWorld-O\cite{yan2025renderworld}  & Occ & None & Pred. &
- & 28.69 & 18.89 & 14.83 & \cellcolor{gray!20}20.80 &
- & 37.74 & 28.41 & 24.08 & \cellcolor{gray!20}30.08 \\

DTT-O\cite{xu2025delta} & Occ & None & Pred. &
85.50 & 37.69 & 29.77 & 25.10 & \cellcolor{gray!20}30.85 &
92.07 & 76.60 & 74.44 & 72.71 & \cellcolor{gray!20}74.58 \\

DFIT-OccWorld-O\cite{zhang2024efficient}  & Occ & None & Pred. &
- & 31.68 & 21.29 & 15.18 & \cellcolor{gray!20}22.71 &
- & 40.28 & 31.24 & 25.29 & \cellcolor{gray!20}32.27 \\

COME-O\cite{shi2025come}   & Occ & None & Pred. &
- & 30.57 & 19.91 & 13.38 & \cellcolor{gray!20}21.29 &
- & 36.96 & 28.26 & 21.86 & \cellcolor{gray!20}29.03 \\

DOME-O\cite{gu2024dome} & Occ & None & GT &
83.08 & 35.11 & 25.89 & 20.29 & \cellcolor{gray!20}27.10 &
77.25 & 43.99 & 35.36 & 29.74 & \cellcolor{gray!20}36.36 \\

COME-O\cite{shi2025come}  & Occ & None & GT &
- & 42.75 & 32.97 & 26.98 & \cellcolor{gray!20}34.23 &
- & 50.57 & 43.47 & 38.36 & \cellcolor{gray!20}44.13 \\

\specialrule{.5pt}{1pt}{1pt}

OccWorld-T\cite{zheng2024occworld} & Camera & Semantic LiDAR & Pred. &
7.21 & 4.68 & 3.36 & 2.63 & \cellcolor{gray!20}3.56 &
10.66 & 9.32 & 8.23 & 7.47 & \cellcolor{gray!20}8.34 \\

OccWorld-S\cite{zheng2024occworld} & Camera & None & Pred. &
0.27 & 0.28 & 0.26 & 0.24 & \cellcolor{gray!20}0.26 &
4.32 & 5.05 & 5.01 & 4.95 & \cellcolor{gray!20}5.00 \\

RenderWorld-S\cite{yan2025renderworld} & Camera & None & Pred. &
- & 2.83 & 2.55 & 2.37 & \cellcolor{gray!20}2.58 &
- & 14.61 & 13.61 & 12.98 & \cellcolor{gray!20}13.73 \\

COME-S\cite{shi2025come} & Camera & None & Pred. &
- & 25.57 & 18.35 & 13.41 & \cellcolor{gray!20}19.11 &
- & 45.36 & 37.06 & 30.46 & \cellcolor{gray!20}37.63 \\

OccWorld-D\cite{zheng2024occworld} & Camera & Occ & Pred. &
18.63 & 11.55 & 8.10 & 6.22 & \cellcolor{gray!20}8.62 &
22.88 & 18.90 & 16.26 & 14.43 & \cellcolor{gray!20}16.53 \\

OccWorld-F\cite{zheng2024occworld} & Camera & Occ & Pred. &
20.09 & 8.03 & 6.91 & 3.54 & \cellcolor{gray!20}6.16 &
35.61 & 23.62 & 18.13 & 15.22 & \cellcolor{gray!20}18.99 \\

OccLLaMA-F\cite{wei2024occllama} & Camera & Occ & Pred. &
37.38 & 10.34 & 8.66 & 6.98 & \cellcolor{gray!20}8.66 &
38.92 & 25.81 & 23.19 & 19.97 & \cellcolor{gray!20}22.99 \\

DFIT-OccWorld-F\cite{zhang2024efficient} & Camera & Occ & Pred. &
- & 13.38 & 10.16 & 7.96 & \cellcolor{gray!20}10.50 &
- & 19.18 & 16.85 & 15.02 & \cellcolor{gray!20}17.02 \\

DTT-F\cite{xu2025delta} & Camera & Occ & Pred. &
43.52 & 24.87 & 18.30 & 15.63 & \cellcolor{gray!20}19.60 &
54.31 & 38.98 & 37.45 & 31.89 & \cellcolor{gray!20}36.11 \\

DOME-F\cite{gu2024dome} & Camera & Occ & GT &
75.00 & 24.12 & 17.41 & 13.24 & \cellcolor{gray!20}18.25 &
74.31 & 35.18 & 27.90 & 23.44 & \cellcolor{gray!20}28.84 \\

COME-F\cite{shi2025come} & Camera & Occ & GT &
- & 26.56 & 21.73 & 18.49 & \cellcolor{gray!20}22.26 &
- & 48.08 & 43.84 & 40.28 & \cellcolor{gray!20}44.07 \\

\specialrule{1pt}{1pt}{0pt}
\end{tabular}%
}
\begin{tablenotes}[flushleft]\fontsize{6pt}{7pt}\selectfont
    \item \hypertarget{note:4DOcc}{}\textsuperscript{1} \textbf{Note}: Method variants are denoted by their input source: O for ground-truth; camera-based variants include D (TPVFormer), F (FBOCC), T (TPVFormer with semantic LiDAR), and S (self-supervised TPVFormer).

    \item \hypertarget{note:Copy_Paste}{}\textsuperscript{2} \textbf{Copy \& Paste}: A naive baseline that repeats the final input frame for all future predictions.
\end{tablenotes}
\end{table*}

\subsection{Control Tasks}\label{subsec:eperformance_control}
\textbf{Evaluation on DMC.}\quad Most studies probe the capacity of world models to learn control-relevant dynamics, typically adopting a pixel-based setting with $64{\times}64{\times}3$ observations. The primary metric is \emph{Episode Return}, defined as the cumulative reward over \num{1000} steps, with a theoretical maximum of \num{1000} given $r_t \in [0,1]$. For comparability, \tabref{tab:DMC} reports the step budget and summarizes performance by task score and task count. The results indicate improved data efficiency, with recent models reaching strong performance in far fewer training steps. However, inconsistent evaluation protocols and task subsets impede a fair assessment of generalization, and building a broadly transferable model across tasks, modalities, and datasets remains an open challenge.

\begin{table}[t]
\centering
\caption{\textbf{Performance comparison on the DMC benchmark\textsuperscript{\protect\hyperlink{note:DMC}{1}}.}}
\label{tab:DMC}
\renewcommand{\arraystretch}{1.2}
\setlength{\tabcolsep}{1.mm}
\resizebox{\columnwidth}{!}{%
\begin{tabular}{l||c|cccc|c}
\specialrule{1.5pt}{0pt}{0pt}
\rowcolor{gray!15}
 & &
\multicolumn{4}{c|}{\textbf{Episode Return}$\uparrow$} & \\
\rowcolor{gray!15}
\multirow{-2}{*}{\textbf{Method}} &
\multirow{-2}{*}{\textbf{Step}} &
\textbf{Reacher Easy} & \textbf{Cheetah Run} & \textbf{Finger Spin} & \textbf{Walker Walk} &
\multirow{-2}{*}{\textbf{Avg. / Total}} \\

\specialrule{0.5pt}{0pt}{1pt}\specialrule{0.5pt}{1pt}{1pt}
PlaNet\cite{hafner2019learning}         & 5M    & 469 & 496 & 495 & 945 & 333/20 \\
Dreamer\cite{hafner2020dream}           & 5M    & 935 & 895 & 499 & 962 & 823/20 \\
Dreaming\cite{okada2021dreaming}        & 500k  & 905 & 566 & 762 & 469 & 610/12 \\
TransDreamer\cite{chen2022transdreamer} & 2M    & -   & \underline{865} & -   & \underline{933} & \underline{893}/4 \\
DreamerPro\cite{deng2022dreamerpro}     & 1M    & 873 & 897 & 811 & -   & 857/6 \\
MWM\cite{seo2023masked}                 & 1M    & -   & \underline{670} &  & -   & \underline{690}/7 \\
HRSSM\cite{sun2024learning}             & 500k  & \underline{910} & -   & \underline{960} & -   & \underline{938}/3 \\
DisWM\cite{wang2025disentangled}        & 1M    & \underline{960} & \underline{820} & -   & \underline{920} & \underline{879}/5 \\
\specialrule{1.5pt}{1pt}{0pt}
\end{tabular}}
\begin{tablenotes}[flushleft]\fontsize{6pt}{7pt}\selectfont
    \item \hypertarget{note:DMC}{}\textsuperscript{1} \textbf{Note}: Performance comparison on the DMC. Underlined entries indicate scores approximated from their respective reward curves. Average scores (Avg.) are provided as a coarse indicator, given the varying difficulty across tasks.

\end{tablenotes}
\end{table}

\para{Evaluation on RLBench.} RLBench evaluates manipulation with a 7-DoF simulated Franka arm and is widely used to assess whether world models capture task-relevant dynamics and support conditioned action generation. The primary metric is \emph{Success Rate}, defined as the fraction of episodes that reach the goal within the step limit. As summarized in \tabref{tab:RLBench}, implementations differ in episode budgets, resolution, and modalities, which complicates like-for-like comparison. Despite this heterogeneity, several trends are evident. Recent methods increasingly leverage multimodal inputs and adopt stronger backbones such as 3DGS and DiT. VidMan achieves a high average success rate on the broadest task, revealing IDM as a promising architectural direction.

\begin{table}[t]
\centering
\caption{\textbf{Performance comparison for manipulation tasks on RLBench.}}
\label{tab:RLBench}
\renewcommand{\arraystretch}{1.2}
\setlength{\tabcolsep}{1.mm}
\resizebox{\columnwidth}{!}{%
\begin{tabular}{cl||ccccc}
\specialrule{1.5pt}{0pt}{0pt}
\rowcolor{gray!15}
\multicolumn{2}{c||}{} & \multicolumn{5}{c}{\textbf{Methods}} \\
\rowcolor{gray!15}
 \multicolumn{2}{c||}{\multirow{-2}{*}{\textbf{Criteria}}}& 
\textbf{VidMan}\cite{wen2024vidman} & \textbf{ManiGaussian}\cite{lu2024manigaussian} & \textbf{ManiGaussian++}\cite{yu2025manigaussian++} & \textbf{DreMa}\cite{barcellona2025dream} & \textbf{TesserAct}\cite{zhen2025tesseract} \\
\specialrule{0.5pt}{0pt}{1pt}\specialrule{0.5pt}{1pt}{1pt}

\parbox[t]{2mm}{\multirow{6}{*}{\rot{\textbf{Setting}}}} &
Episode & 125 & 25 & 25 & 250 & 100 \\
& Pixel & 224 & 128 & 256 & 128 & 512 \\
& Depth &  & \cmark & \cmark & \cmark & \cmark \\
& Language & \cmark & \cmark & \cmark & & \cmark \\
& Proprioception & \cmark & \cmark & \cmark & & \\
& Characteristic & IDM & GS & Bimanual & GS & DiT \\
\cmidrule{2-7}

\parbox[t]{2mm}{\multirow{6}{*}{\rot{\textbf{Success Rate (\%)}}}} & 
Stack Blocks & 48 & 12 & - & 12 & - \\
& Close Jar & 88 & 28 & - & 51 & 44 \\
& Open Drawer & 94 & 76 & - & - & 80 \\
& Sweep to Dustpan & 93 & 64 & 92 & - & 56 \\
& Slide Block & 98 & 24 & - & 62 & - \\
\cmidrule{2-7}

& \textbf{Avg.\textsuperscript{\protect\hyperlink{note:RLBench}{1}} / Total} & 67/18 & 45/10 & 35/10 & 25/9 & 63/10 \\
\specialrule{1.5pt}{1pt}{0pt}
\end{tabular}}
\begin{tablenotes}[flushleft]\fontsize{6pt}{7pt}\selectfont
    \item \hypertarget{note:RLBench}{}\textsuperscript{1} \textbf{Avg.}: Average scores are reported only as a coarse indicator, given varying task difficulty.
\end{tablenotes}
\end{table}

\para{Planning on nuScenes.} Open-loop planning is treated as a world modeling task on the nuScenes validation split, where models predict ego motion from a limited history. Methods observe \SI{2}{\second} of past trajectories and forecast the next \SI{3}{\second} as 2D BEV waypoints. Evaluation reports \emph{L2} error and \emph{collision rate} at multiple horizons, and \tabref{tab:nuscenes_planning} summarizes results by input modality, auxiliary supervision, and metric settings. Under this shared protocol, a clear tradeoff emerges. UniAD+DriveWorld achieves the lowest \emph{L2} with extensive auxiliary supervision, whereas SSR attains the best collision rate with competitive \emph{L2} without extra supervision. Camera-based methods now surpass models that use privileged occupancy, reflecting the growing maturity of E2E planning.

\begin{table*}[t]
\centering
\caption{\textbf{Performance comparison for open-loop planning on the nuScenes validation split\textsuperscript{\protect\hyperlink{note:Planning}{1}}.}}
\label{tab:nuscenes_planning}
\renewcommand{\arraystretch}{1.1}
\resizebox{\linewidth}{!}{%
\begin{tabular}{l||cc||C{.05\textwidth}C{.05\textwidth}C{.05\textwidth}C{.05\textwidth}C{.05\textwidth}C{.05\textwidth}C{.05\textwidth}C{.05\textwidth}}
\specialrule{1.5pt}{0pt}{0pt}
 & \multirow{2}{*}{\textbf{Input}} & \multirow{2}{*}{\textbf{Aux. Sup.\textsuperscript{\protect\hyperlink{note:aux_sup}{2}}}} & \multicolumn{4}{c}{\textbf{L2 (m)} $\downarrow$} & \multicolumn{4}{c}{\textbf{Collision Rate (\%) $\downarrow$}} \\
\multirow{-2}{*}{\textbf{Method}} & & & \textbf{1s} & \textbf{2s} & \textbf{3s} & \cellcolor{gray!20}\textbf{Avg.} & \textbf{1s} & \textbf{2s} & \textbf{3s} & \cellcolor{gray!20}\textbf{Avg.} \\
\specialrule{0.5pt}{.5pt}{0.5pt}\specialrule{0.5pt}{0.5pt}{.5pt}
UniAD\cite{hu2023planning} & Camera & Map \& Box \& Motion \& Tracklets \& Occ 
& 0.48 & 0.96 & 1.65 & \cellcolor{gray!20}1.03 
& 0.05 & 0.17 & 0.71 & \cellcolor{gray!20}0.31 \\

UniAD+DriveWorld\cite{min2024driveworld} & Camera & Map \& Box \& Motion \& Tracklets \& Occ
& 0.34 & 0.67 & 1.07 & \cellcolor{gray!20}0.69
& 0.04 & 0.12 & 0.41 & \cellcolor{gray!20}0.19 \\

GenAD\cite{zheng2024genad} & Camera & Map \& Box \& Motion 
& 0.36 & 0.83 & 1.55 & \cellcolor{gray!20}0.91 
& 0.06 & 0.23 & 1.00 & \cellcolor{gray!20}0.43 \\

FSDrive\cite{zeng2025futuresightdrive} & Camera &  Map \& Box \& QA
& 0.40 & 0.89 & 1.60 & \cellcolor{gray!20}0.96
& 0.07 & 0.12 & 1.02 & \cellcolor{gray!20}0.40 \\

OccWorld-T\cite{zheng2024occworld} & Camera & Semantic LiDAR
& 0.54 & 1.36 & 2.66 & \cellcolor{gray!20}1.52
& 0.12 & 0.40 & 1.59 & \cellcolor{gray!20}0.70 \\

Doe-1\cite{zheng2024doe} & Camera & QA 
& 0.50 & 1.18 & 2.11 & \cellcolor{gray!20}1.26
& 0.04 & 0.37 & 1.19 & \cellcolor{gray!20}0.53 \\

SSR\cite{li2025navigation} & Camera & None 
& 0.24 & 0.65 & 1.36 & \cellcolor{gray!20}0.75 
& 0.00 & 0.10 & 0.36 & \cellcolor{gray!20}0.15 \\

OccWorld-S\cite{zheng2024occworld} & Camera & None
& 0.67 & 1.69 & 3.13 & \cellcolor{gray!20}1.83
& 0.19 & 1.28 & 4.59 & \cellcolor{gray!20}2.02 \\

Epona\cite{zhang2025epona} & Camera & None
& 0.61 & 1.17 & 1.98 & \cellcolor{gray!20}1.25
& 0.01 & 0.22 & 0.85 & \cellcolor{gray!20}0.36 \\

RenderWorld\cite{yan2025renderworld} & Camera & None
& 0.48 & 1.30 & 2.67 & \cellcolor{gray!20}1.48
& 0.14 & 0.55 & 2.23 & \cellcolor{gray!20}0.97 \\

Drive-OccWorld\cite{yang2025driving} & Camera & None
& 0.32 & 0.75 & 1.49 & \cellcolor{gray!20}0.85
& 0.05 & 0.17 & 0.64 & \cellcolor{gray!20}0.29 \\

OccWorld-D\cite{zheng2024occworld} & Camera & Occ
& 0.52 & 1.27 & 2.41 & \cellcolor{gray!20}1.40
& 0.12 & 0.40 & 2.08 & \cellcolor{gray!20}0.87 \\

OccWorld-F\cite{zheng2024occworld} & Camera & Occ
& 0.45 & 1.33 & 2.25 & \cellcolor{gray!20}1.34
& 0.08 & 0.42 & 1.71 & \cellcolor{gray!20}0.73 \\

OccLLaMA-F\cite{wei2024occllama} & Camera & Occ
& 0.38 & 1.07 & 2.15 & \cellcolor{gray!20}1.20
& 0.06 & 0.39 & 1.65 & \cellcolor{gray!20}0.70 \\

DTT-F\cite{xu2025delta}& Camera & Occ
& 0.35 & 1.01 & 1.89 & \cellcolor{gray!20}1.08
& 0.08 & 0.33 & 0.91 & \cellcolor{gray!20}0.44 \\ 

DFIT-OccWorld-V\cite{zhang2024efficient} & Camera & Occ
& 0.42 & 1.14 & 2.19 & \cellcolor{gray!20}1.25
& 0.09 & 0.19 & 1.37 & \cellcolor{gray!20}0.55 \\

NeMo\cite{huang2024neural} & Camera & Occ
& 0.39 & 0.74 & 1.39 & \cellcolor{gray!20}0.84
& 0.00 & 0.09 & 0.82 & \cellcolor{gray!20}0.30 \\

\specialrule{.5pt}{1pt}{1pt}

OccWorld-O\cite{zheng2024occworld} & Occ & None
& 0.43 & 1.08 & 1.99 & \cellcolor{gray!20}1.17
& 0.07 & 0.38 & 1.35 & \cellcolor{gray!20}0.60 \\
 
OccLLaMA-O\cite{wei2024occllama} & Occ & None
& 0.37 & 1.02 & 2.03 & \cellcolor{gray!20}1.14
& 0.04 & 0.24 & 1.20 & \cellcolor{gray!20}0.49 \\

RenderWorld-O\cite{yan2025renderworld} & Occ & None
& 0.35 & 0.91 & 1.84 & \cellcolor{gray!20}1.03
& 0.05 & 0.40 & 1.39 & \cellcolor{gray!20}0.61 \\ 

DTT-O\cite{xu2025delta}& Occ & None
& 0.32 & 0.91 & 1.76 & \cellcolor{gray!20}1.00
& 0.08 & 0.32 & 0.51 & \cellcolor{gray!20}0.30 \\ 

DFIT-OccWorld-O\cite{zhang2024efficient} & Occ & None
& 0.38 & 0.96 & 1.73 & \cellcolor{gray!20}1.02
& 0.07 & 0.39 & 0.90 & \cellcolor{gray!20}0.45 \\ 
\specialrule{1pt}{1pt}{0pt}
\end{tabular}%
}
\begin{tablenotes}[flushleft]\fontsize{6pt}{7pt}\selectfont
    \item \hypertarget{note:Planning}{}\textsuperscript{1} \textbf{Note}: Results are reported following the UniAD~\cite{hu2023planning} protocol. Method variants are denoted by their input source: O (ground-truth), V (camera-predicted), D (TPVFormer), F (FBOCC), \etc.
    \item \hypertarget{note:aux_sup}{}\textsuperscript{2} \textbf{Aux. Sup.}: Abbreviation for auxiliary supervision, which refers to additional training signals beyond ground-truth trajectories.
\end{tablenotes}
\end{table*}

\section{Challenges and Trends}
\label{sec:challenges}
This section reviews the open challenges and emerging directions for world models in embodied AI. We organize them across three dimensions: \secref{subsec:data} Data \& Evaluation, \secref{subsec:computational} Computational Efficiency, and \secref{subsec:modeling} Modeling Strategies.

\subsection{Data \& Evaluation}\label{subsec:data}
\textbf{Challenges.}\quad From a data perspective, the central challenge lies in the scarcity and heterogeneity of existing corpora. Although embodied AI spans diverse domains such as navigation, manipulation, and autonomous driving, a unified large-scale dataset remains lacking. This fragmentation constrains the capacity of world models and substantially hinders their ability to generalize, especially when model scale grows faster than available task-relevant data. Multi-agent interaction data are particularly underrepresented, limiting progress on socially aware and cooperative world models.

Evaluation practices face similar limitations. Most metrics, such as FID and FVD emphasize pixel fidelity while ignoring  physical consistency, dynamics, and causality. As a result, models may excel at appearance-level realism yet exhibit physically implausible or causally inconsistent rollouts. Existing benchmarks (\eg, EWM-Bench~\cite{yue2025ewmbench}) begin to introduce more structured metrics, but remain task-specific and lack cross-domain standards, and there is still a gap between how video simulators (\eg, Sora-style models) are evaluated and how embodied controllers are judged.

\para{Future Directions.} 
Recent initiatives such as OpenDV-2K~\cite{yang2024generalized}, VideoMix22M~\cite{assran2025v}, and VisionMix163M~\cite{mur2026v} highlight the growing focus on large-scale pretraining and broader modality coverage, yet resources remain fragmented and domain specific. A key direction is to construct unified, multimodal, cross-domain benchmarks that couple scenes, actions, and rewards, and that explicitly cover multi-agent interaction and safety-critical events. In parallel, evaluation protocols should move beyond perceptual realism toward metrics that assess physical consistency, causal reasoning, and long-horizon error accumulation. Bridging the evaluation gap between video-centric world simulators and embodied action benchmarks is essential if large video models are to serve as reliable substrates for downstream control.

\subsection{Computational Efficiency}
\label{subsec:computational}
\textbf{Challenges.}\quad Embodied AI tasks encounter significant challenges in computational efficiency, particularly in real-time settings. Although Transformers and diffusion-based decoders achieve strong performance, their high inference costs and memory footprint conflict with the latency and energy constraints of onboard robotic platforms. This tension is exacerbated by the mismatch between rapidly growing model sizes and relatively modest compute budgets on edge devices. As a result, many practical systems still rely on compact recurrent models and global latent vectors, accepting limited capacity in exchange for predictable real-time behavior.

\para{Future Directions.} To address this challenge, future research should emphasize architectures and training schemes that jointly optimize fidelity and efficiency. Model compression techniques such as quantization, pruning, low-rank and sparse computation are natural candidates to reduce latency and memory while preserving control-relevant performance. Sequence models such as SSMs offer linear-time scaling and show promise for long-horizon reasoning under tight compute budgets. More broadly, it may be beneficial to design world models as adaptive systems that can adjust resolution, rollout horizon, or modality usage to available resources. For example, high-capacity
video-diffusion components may be used primarily for offline data generation or simulation, while lightweight cores are reserved for online control and safety monitoring.

\subsection{Modeling Strategy}\label{subsec:modeling}
\textbf{Challenges.}\quad Despite rapid progress, world models still struggle with long-horizon temporal dynamics and efficient spatial representations. A central difficulty is balancing recurrent simulation and global prediction: autoregressive designs are compact and sample-efficient but suffer from error accumulation, whereas global prediction improves multi-step coherence at the cost of heavy computation and weaker closed-loop interactivity. 
On the spatial side, efficiency and inductive bias remain bottlenecks. Latent vectors, token sequences, spatial grids, and decomposed rendering representations each exhibit distinct trade-offs between expressiveness, geometric fidelity, and computational cost. Highly realistic video or 3D generative models often capture correlations in appearance without robust causal structure, leading to visually plausible yet physically incorrect predictions.

\para{Future Directions.} Several promising avenues have emerged to address these bottlenecks. SSM-based architectures (\eg, Mamba) align naturally with autoregressive modeling and offer linear-time scalability, while masked prediction frameworks (\eg, JEPA-style objectives) support global, parallel prediction and stronger representation learning. Hybrid designs that combine local autoregression with global or masked updates, possibly supported by explicit memory modules or hierarchical planning, could reduce error accumulation and improve long-horizon stability.

On the spatial axis, future work should seek representations that encode geometry and physics more explicitly while remaining tractable for control. For example, hybrid token-grid or neural field-based models can expose causal structure and uncertainty to downstream policies. Looking ahead, a unifying perspective is to view the world model as shared infrastructure that coordinates perception, prediction, and decision-making for embodied agents. Realizing this perspective will require architectures that tightly integrate temporal and spatial modeling, scale effectively with model and data size, and support principled multi-agent, multi-task, and multi-modal interactions.

\section{Conclusion}\label{sec:conclusion}
This survey has organized world models for embodied AI within a three-axis framework spanning decision coupling, temporal modeling, and spatial representation. Using this lens, we mapped methods in robotics, autonomous driving, and general-purpose video modeling into a unified design space, summarized representative architectures and
benchmarks, and analyzed how different combinations of these axes give rise to distinct trade-offs in fidelity, efficiency, and interactivity.

Looking ahead, progress in embodied world models is likely to be driven by three converging trends: scaling general-purpose video and 3D world models, tightening their coupling to control through action- and reward-conditioned training, and improving robustness via causally and physically grounded objectives. Rather than a single unified benchmark, we expect families of cross-domain, multimodal suites that jointly assess prediction quality, decision making, and safety under sim-to-real shift. Although world models are unlikely to close the simulation-reality gap entirely, advances in geometry, physics, and uncertainty modeling should make simulated rollouts more informative for downstream planning and policy learning, positioning world models as shared predictive infrastructure within embodied AI systems.



{\small
\bibliographystyle{IEEEtran}
\bibliography{reference}
}

\vfill


\end{document}